# Title: Enhancing Prediction and Analysis of UK Road Traffic Accident Severity Using AI: Integration of Machine Learning, Econometric Techniques, and Time Series Forecasting in Public Health Research

Author Name: Md Abu Sufian[1], Jayasree Varadarajan[2]

Affiliation: University of Leicester, University Rd, Leicester LE1 7RH, United Kingdom

Email: Abusufian.tex.cu@gmail.com[1] & jayasreevaradhan@gmail.com

# Declaration of Interest Statement:

We declare that there are no conflicts of interest regarding the submission of our paper Title: "Enhancing Prediction and Analysis of UK Road Traffic Accident Severity Using AI: Integration of Machine Learning, Econometric Techniques, and Time Series Forecasting in Public Health Research" to the respected journal ".

Ethics approval and consent to participate:
Not applicable. The study involved the analysis of publicly available data with no identifiable human information.

Consent for publication:
Not applicable. This manuscript does not contain any individual person's data.

Availability of data and materials:
The data that support the findings of this study are available from the UK Department for Transport but restrictions apply to the availability of these data, which were used under license for this study. Data are however available from the authors upon reasonable request and with permission of the UK Department for Transport.

Competing interests:
The authors declare that they have no competing interests.

Funding:
No external funding was received for this research.

Authors' contributions:
First author, Md Abu Sufian performed data cleaning, statistical analysis, model building and second author, Jayasree Varadarajan contributed to literature review and entire paper review and further correction entire paper to interpretation under the guidance of the supervisory team. Md Abu Sufian drafted the manuscript and all authors contributed to critical revisions of the paper and approved the final version for submission.

Acknowledgements:
We confirm that this manuscript represents original work and has not been published previously or submitted elsewhere for consideration. We are committed to full transparency and accountability in the research process, and we assure the journal of the integrity and validity of our study. We would like to acknowledge the UK Department for Transport for making the data used in this study available. and University of Leicester Mathematical and computer science department and corresponded supervisor professor kings Paul, Data Science. We would also like to thank our peers for their valuable insights and suggestions during the drafting of this manuscript.

We understand that the journal's decision on the publication of our paper will be based solely on its scientific merit.

Thank you for considering our manuscript.

Sincerely,

Md Abu Sufian
University of Leicester
Leicester, United Kingdom
abusufian.tex.cu@gmail.com

# 1 Abstract

This research project delves into the intricacies of road traffic accidents severity in the UK, employing a potent combination of machine learning algorithms, econometric techniques, and traditional statistical methods to analyse longitudinal historical data. Our robust analysis framework includes descriptive, inferential, bivariate, and multivariate methodologies, correlation analysis: Pearson's and Spearman's Rank Correlation Coefficient, multiple and logistic regression models, Multicollinearity Assessment, and Model Validation. In addressing heteroscedasticity or autocorrelation in error terms, we've advanced the precision and reliability of our regression analyses using the Generalized Method of Moments (GMM). Additionally, our application of the Vector Autoregressive (VAR) model and the Autoregressive Integrated Moving Average (ARIMA) models have enabled accurate time-series forecasting. With this approach, we've achieved superior predictive accuracy, marked by a Mean Absolute Scaled Error (MASE) of 0.800 and a Mean Error (ME) of -73.80 compared to a naive forecast. The project further extends its machine learning application by creating a random forest classifier model with a precision of 73 per cent, a recall of 78 per cent, and an F1-score of 73 per cent. Building on this, we employed the H2O AutoML process to optimize our model selection, resulting in an XGBoost model that exhibits exceptional predictive power, as evidenced by an RMSE of 0.1761205782994506 and MAE of 0.0874235576229789. Factor Analysis was leveraged to identify underlying variables or factors that explain the pattern of correlations within a set of observed variables. Scoring history, a tool to observe the model's performance throughout the training process, was incorporated to ensure the highest possible performance of our machine learning models. We also incorporated Explainable AI (XAI) techniques, utilizing the SHAP (Shapley Additive Explanations) model to comprehend the contributing factors to accident severity. Features such as Driver_Home_Area_Type, Longitude, Driver_IMD_Decile, Road_Type, Casualty_Home_Area_Type, and Casualty_IMD_Decile were identified as significant influencers. Our research contributes to the nuanced understanding of traffic accident severity and demonstrates the potential of advanced statistical, econometric and machine learning techniques in informing evidence-based interventions and policies for enhancing road safety.

**Keywords:** Statistical Analysis, Machine Learning, ARIMA, Winsorization, Explainable AI (XAI), SHAP (Shapley Additive Explanations), Public Health, Road Safety, Policy, GMM, VAR, Factor Analysis, H2O AutoML, XGBoost, Random Forest Classifier.

# 2 Introduction

## 2.1 Background

Road traffic accidents are a major public health concern and injuries worldwide. In 2019, 1,752 people died and 25,945 people suffered serious injuries in road traffic incidents in the UK **[29]**. For successful interventions and policies to reduce the number of accidents and enhance road safety, it is essential to understand the elements that contribute to the severity of these accidents **[6]**. Additionally, during the period 2010-2019, there was a 30% decline in the number of people killed in road traffic accidents. However, the number of major and minor injuries has remained relatively stable throughout this period. The main aim of this project is to investigate the variables that affect the UK's serious incidents of accidents on roads. To conduct this study of 2019 road safety statistics, R programming language will be employed.

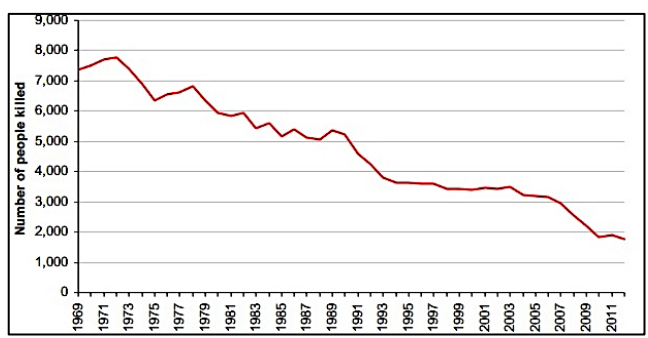

**Fig. 1** Death Cases Reported **[2]**

According to **Fig. 1**, the number of people killed due to road accidents has reduced drastically. Possible reasons may be due to the advanced technology, road safety and education and road policies being implemented. According to the study conducted by Xu **[15]**, the results indicated that several factors, such as driving behaviour, car characteristics, road infrastructure, and environmental conditions affect the severity of traffic road accidents. Yet, there is currently a lack of studies that have used rigorous statistical analysis methods to study the multivariate drivers of the severity of road traffic accidents in the UK. Age, gender, and location are just a few examples of factors that can affect how likely you are to be in a traffic accident. Young people, who make up only 7% of the population but 20% of all fatalities in road traffic incidents in 2019, disproportionately feel the impact. Road traffic accidents are more likely to involve men than women, and they happen more frequently in cities than in rural areas.

## 2.2 Importance of the study

Road traffic accidents contribute to more than 1.3 million deaths and 50 million injuries each year, making it one of the leading sources of death and injury globally. As a result of these accidents, there is a large loss of human life, property damage, and significant healthcare costs **[9]**. The research on road traffic accident severity is therefore important since it helps in identifying the causes of accidents and the development of effective measures to minimize the frequency and severity. Analysing various elements that would influence an accident's severity is a necessary step in studying road traffic accident severity. These variables fall into three categories: driver, vehicle, and road-related variables **[1]**. Driver behaviour, such as speeding, driving while drunk, and distracted driving are examples of factors associated with drivers. The type and condition of the vehicle involved in the accident, as well as its age, safety features, and maintenance history, may be considered vehicle-related variables. Road design, traffic volume, and weather conditions are illustrations of factors that may be associated with roads **[2].** Understanding the causes of traffic accidents and the best treatments to minimize the severity can be achieved by analysing these elements. For instance, the development of specialized education and awareness programs to encourage safe driving habits may result from the understanding of driver behaviour as a significant determinant in accident severity. Similarly, to that, if bad road design is found to be a contributory factor, adjustments to the road infrastructure, such as the addition of safety barriers or altered road markings, may be made. Analysing the severity of traffic accidents can also assist in deciding which treatments to prioritize and how to spend resources. The elements that have the greatest impact on the severity of accidents must be prioritized using the given limited resources available to address road safety. Policymakers and organizations that promote road safety can establish the most efficient initiatives and spend resources accordingly by examining the data on accident severity. Studying the severity of traffic accidents can also be useful in assessing the effectiveness of road safety initiatives. Following the implementation of road accident measures, it is important to assess the effectiveness by lowering the accident severity. Accident severity research can be used to assess the success of initiatives and point out areas that still require attention. The analysis of accident severity data, for instance, can be used to assess if a speeding-related accident reduction program has been successful in reducing the number of accidents. The creation of predictive models for accident severity can also be important in the study of the severity of traffic accidents. Using historical data, predictive models can be implemented to predict the occurrence of the severity of accidents in the future. These models can be used to prioritize interventions and inform planning for improving road safety. Predictive models, for instance, can also be used to pinpoint accident hotspots, enabling the adoption of focused interventions to minimise the severity of accidents. In conclusion, research into the severity of traffic accidents is important since it offers insightful information about their causes and contributes to the development of practical solutions for lowering both their incidence and severity. Policymakers and road safety groups can efficiently prioritize actions, distribute resources, assess the success of interventions, and create predictive models for accident severity by evaluating data on accident severity. The reduction of accidents, injuries, and fatalities on the roads will make roads safer for everyone, and this is the goal of research into road traffic accident severity.

## 2.3 Aim

This study aims to contribute to the existing body of knowledge on road traffic accident severity and inform the development of effective policies and interventions to reduce the human and financial costs associated with accidents in the UK.

## 2.4 Objectives

1. Investigate the relationship between driver age and the severity of road traffic accidents in the UK, with a focus on Casualty Status 19, and identify the age groups that are more likely to be involved in severe accidents.

2. Analyze the seasonal trends in the rate of road traffic accidents in the UK, determine how these trends differ for accidents of varying severity, and identify the periods when more severe accidents are more likely to occur.
3. Assess the impact of various environmental and road conditions, such as lighting, weather, road quality, and road size, on the number of casualties in road traffic accidents in the UK, with a focus on Casualty Status 19.
4. Examine the relationship between the number of vehicles involved in an accident and the severity of the accident and determine how multiple-vehicle accidents contribute to accident severity.
5. Investigate the impact of driver distractions, such as using a mobile phone or eating, on the severity of road traffic accidents in the UK.
6. Provide evidence-based recommendations for policymakers and stakeholders to develop targeted strategies for reducing the frequency and severity of road traffic accidents in the UK, taking into consideration the relationships and trends identified through the study.

## 2.5 Research Questions:

1. With an emphasis on Casualty Status 19, how does driver age relate to the seriousness of traffic incidents in the UK? Are there any age groups that have a higher risk of being in serious accidents?
2 Can seasonal trends be seen in the number of auto accidents in the UK? In that case, how do these patterns change for incidents of varying seriousness? When are accidents of greater severity more likely to happen?
3. What is the impact of various conditions, such as lighting, weather, road quality, and road size, on the number of casualties in road traffic accidents in the UK, with a focus on Casualty Status 19?
4. How does the number of vehicles involved in an accident affect the severity of the accident?
5. What is the impact of driver distraction, such as using a mobile phone or eating, on the severity of road traffic accidents?

## 2.6 Hypothesis:

H1: Drivers who are older than middle-aged people are more likely to be involved in serious accidents due to both these factors and the severity of traffic accidents.
H2: The number of motor vehicle incidents in the UK is affected by seasonal patterns, with more serious accidents occurring in some seasons (like the winter) due to unfavourable weather.
H3a: Traffic accident severity and inadequate lighting are positively correlated in the UK.
H3b: In the UK, severe traffic accidents are positively correlated with adverse weather conditions like rain, snow, and fog.
H3c: Inadequate Road conditions and bigger roads are factors in the UK's rising number and severity of traffic accidents.
H4: The severity of an accident is positively correlated with the number of vehicles involved.
H5: Driver distractions like eating or using a cell phone are positively correlated with the severity of traffic accidents in the UK.

## 3 Literature Review

### 3.1 Theoretical Frameworks

Several theoretical frameworks and concepts connected to road traffic accidents and severity are used to comprehend the dynamic interrelationship of factors that lead to accidents and accident severity. These theoretical frameworks and concepts provide the foundation for assessing traffic accidents and creating successful countermeasures to minimize the incidence and severity. Below theoretical frameworks **Fig. 2**

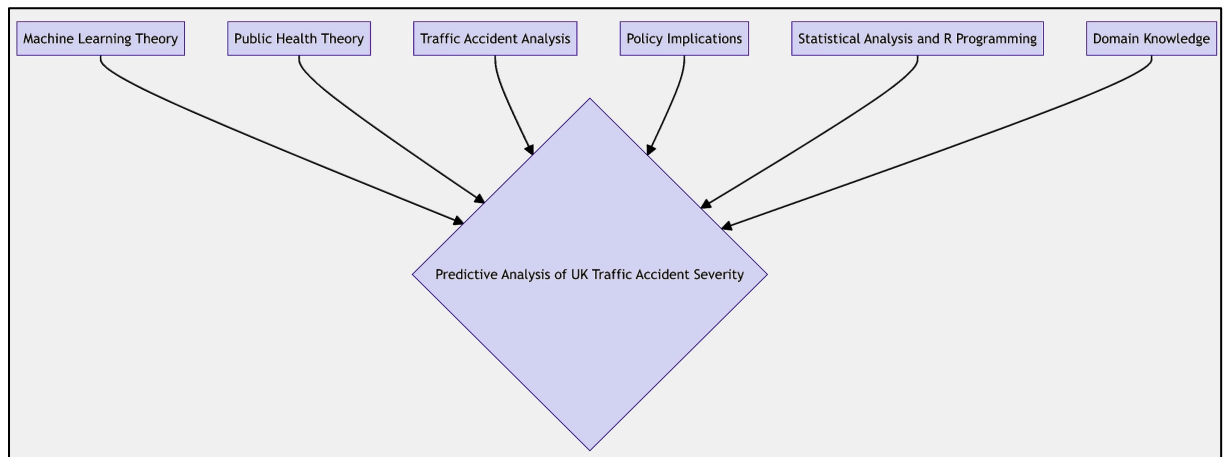

**Fig. 2** Theoretical Framework of Predictive Analysis of UK Traffic Accident Severity

### 3.2 ARIMA (Autoregressive Integrated Moving Average) Model:

ARIMA is a forecasting technique that projects the future values of a series based entirely on its own inertia. Its main application is in short-term forecasting requiring at least 40 historical data points. It works best when your data exhibits a stable or consistent pattern over time with a minimum number of outliers. **[23]**

The ARIMA model is described by three parameters: (p, d, q)

 - p: The order of the autoregressive part (AR). It allows us to incorporate the effect of past values into our model. Essentially, this is the number of lag observations included in the model.
 - d: The order of differencing. It's the number of times the data have had past values subtracted. Differencing is used to make the time series stationary.
 - q: The order of the moving average part (MA). This allows us to set the error of our model as a linear combination of the error values observed at previous time points in the past.

The mathematical model for ARIMA is: (Box, G.E., ... 2015.)

$$Y(t) = C + P_1 \cdot Y(t-1) + P_2 \cdot Y(t-2) + ... + P_p \cdot Y(t-p) + q_1 \cdot e(t-1) + q_2 \cdot e(t-2) + ... + q_n \cdot e(t-q) + e(t)$$

 - $Y(t)$ is the predicted value for the series at time t,
 - C is the constant,
 - $P_1$ through $P_p$ are the parameters of the autoregressive part,
 - $q_1$ through $q_n$ are the parameters of the moving average part,
 - $e(t)$ is the error term of the model.

### 3.3 SHAP (SHAPley Additive Explanations):

SHAP is a game theoretic approach to explain the output of any machine learning model. It connects optimal credit allocation with local explanations using the classic Shapley values from game theory and their related extensions. The SHAP explanation method computes Shapley values for feature importance. Shapley values calculate the importance of a feature by comparing what a model predicts with and without the feature. However, since the order in which a model sees features can affect its predictions, this is done in every possible order, so that the Shapley value is an average of all possible marginal contributions across all permutations (all possible orderings). Mathematically, the Shapley value is the weighted average of the marginal contributions of a feature value to the prediction in different coalitions. **[19][20]**

The Shapley value of feature i is given by:

$$\varphi_i(f) = \sum_{S \subseteq N \setminus \{i\}} \left[ \frac{|S|!(|N|-|S|-1)!}{|N|!} \right] \cdot (f_{S \cup \{i\}}(x) - f_S(x))$$

 - N is the set of all features,
 - S is a subset of N,
 - $f_S(x)$ is the prediction that the model makes with feature set S,
 - $f_{S \cup \{i\}}(x)$ is the prediction that the model makes with feature set S union feature i.

- "!" is the factorial operation

The Shapley value φi(f) is the average contribution of feature i to the prediction for all possible sets of features.

### 3.4 Public Health Theory:

#### 3.4.1 Risk factors for accidents:
Understanding risk factors is a key principle in public health. In the context of traffic safety, risk factors could include driver behaviour (like drunk or distracted driving), environmental factors (such as poor road conditions or bad weather), and demographic factors (like the age and sex of the driver). A review study published in 2017 in "Current Epidemiology Reports" **[24]** found that younger drivers and male drivers tend to have higher rates of fatal accidents. Identifying and understanding these risk factors can help target interventions for those most at risk.

#### 3.4.2 Interventions to prevent accidents:
Once we understand the risk factors, we can design interventions to address them. Interventions might include education campaigns about the dangers of drunk driving, improvements in road design to make them safer, or the development of new vehicle safety features. A review by the Cochrane Collaboration in 2014 found that certain educational interventions can have a positive effect on improving driver behaviour and reducing accidents.**[25]**

#### 3.4.3 The role of policy in promoting public health:
The policy can play a crucial role in promoting public health by implementing regulations that discourage risky behaviour and promote safety. For example, laws that impose penalties for drunk driving can deter people from engaging in this behaviour, thereby reducing the risk of accidents. In a study published in the American Journal of Public Health in 2014,**[26]** it was found that strict enforcement of traffic laws led to a significant reduction in road traffic deaths.

These principles of public health - identifying risk factors, developing interventions, and implementing policy - can be applied to many other areas of public health, not just traffic safety.

#### 3.4.4 Traffic Conflict Techniques (TCT)

Traffic Conflict Techniques are observational methods used to identify and evaluate potential accidents or near-miss events in traffic. These techniques can help identify problem areas in the road network and allow researchers to analyse risk factors contributing to accident severity without relying solely on historical accident data **[4]**. In the UK, this method is frequently employed to assist in determining the causes of accidents and in the development of plans to lower the incidence of fatalities. TCT entails looking at how drivers behaved before an accident and identifying situations when they may have caused a conflict that could have led to an accident according to Sidiq, 2022. Eyewitness testimonials or video recordings of the accident can be used for this analysis. Researchers may determine the major contributing elements to the catastrophe and create plans to remedy them by identifying these clashing time frames. One advantage of employing TCT is that it can assist in figuring out the basic reasons why accidents happen, which might not be easily apparent from a surface-level inspection. For instance, researcher Sidiq said, TCT may show that a specific road design or signage is confusing drivers and increasing the rate of accidents in a specific location. Then, with the use of this data, roads can be made safer and more accident-free.

#### 3.4.5 Risk Compensation Theory

Risk Compensation Theory suggests that drivers adapt their behaviour based on their perceived level of risk. For example, drivers may drive more cautiously in adverse weather conditions or when they are aware of increased police presence **[5]**. This theory highlights the importance of understanding driver behaviour in relation to road safety measures and interventions. According to this idea, drivers may take greater risks when they feel safer, for as when operating a vehicle with advanced safety systems or on a route that has been well maintained, which is relevant to UK traffic incidents. The risk compensation theory's main consequence is that improvements in traffic safety might not necessarily result in a proportionate reduction in the frequency of accidents. For instance, if a new safety device, like anti-lock brakes, is added to cars, drivers might become more assured in their ability to drive safely and take more risks, which would reduce the effectiveness of the safety feature. In the area of traffic safety, this concept has generated a lot of discussion. Some researchers contend that risk compensation is a real occurrence that should be considered when creating strategies for improving road safety, while others contend that the idea is exaggerated and that advances in road safety still result in a net decrease in the number of accidents. Risk compensation theory is still a crucial idea in the UK's field of road safety despite the controversies surrounding it according to Brubacher. It implies that altering driver behaviour and attitudes towards risk is just as important as adding new safety systems to increase road safety. The various factors that affect driver behaviour, such as risk perception, attitudes towards safety, and societal conventions, must be considered for road safety policies to be effective.

### 3.4.6  Road Safety Audit (RSA)

Road Safety Audit is a systematic process for evaluating the safety performance of a road design or infrastructure. This process can identify potential safety hazards and deficiencies, enabling the implementation of corrective measures to improve overall road safety and reduce accident severity **[6]**. The inclusion of this concept could help emphasize the importance of road design and infrastructure in accident prevention. RSA normally entails a team of experts in road safety who carry out a thorough analysis of a road or highway project, from design to construction and operation **[6]**. The team looks at a variety of elements that may have an impact on road safety, such as road geometry, traffic flow, signage and markings, surface conditions, and the actions of other road users. The RSA team evaluates potential security risks and suggests workable security solutions to deal with them. According to the AlHamad researcher, these actions could involve rearranging the way the road is laid out, making modifications to the signage and markings, taking steps to slow down traffic, or adding additional safety elements like crash barriers or pedestrian crossings. Due to its ability to identify and address safety concerns before they cause accidents, RSA is a crucial instrument for enhancing road safety in the UK. RSA can assist prevent accidents from happening in the first place rather than just lessening their effects by including safety considerations in the planning and design phases of a road project.

### 3.4.7  The Haddon Matrix

A theoretical framework for analysing the causes and seriousness of accidents is the Haddon matrix. Please see **Table 1** for more details. According to the research done by researcher Garkaz,2020, the parameters influencing the severity of injuries brought on by traffic accidents were identified using the Haddon matrix. The participants' average age was 33.63 ± 18.53 years. Males between the ages of 17 and 30 made up most of the victims of road accidents and suffered severe and critical injuries. Car-pedestrian crashes (27.9%), vehicle overturns (31.1%), and collisions between two cars (26.3%) were the most frequent causes of injury **[12]**. The most frequent moving offences were exceeding the speed limit (73.2%) and failing to provide the right-of-way (17.9%). A strong correlation between the accident's time and place and the severity of the injuries was also found, according to the results of the multivariate analysis of data from car safety devices (p 0.001) **[12].**

| Characteristics | Category | Frequency (%) |
|---|---|---|
| Age ( Years) | 001-16 | 323 (16) |
| | 17-29 | 694 (34.4) |
| | 30-39 | 362 (18) |
| | 40-49 | 262 (13) |
| | 50-65 | 262(12.3) |
| | >65 | 127(6.3) |
| Gender | Man | 1474 (73.2) |
| | Woman | 541.(26.8) |
| Type of Accident | Pesdestrian with car | 563 ( 27.9) |
| | Motorcycle with car | 232(11.5) |
| | Overturning | 626(31.1) |
| | Two Car Collision | 530(26.30) |
| | Two Motorcycle Collision | 22(1.1) |
| | Motorcycle with Pedestrian | 31(1.5) |
| Accident Location | Inside the City | 880(43.7) |
| | Outside the city | 1133(56.2) |
| Time of Accident ( Hour) | 8 am-2 am | 659(32.7) |
| | 3am -8 pm | 797(39.6) |
| | 21pm-8am | 559(27.7) |
| Injury Score* | Mild(1-9) | 800(39.7) |
| | Moderate(10-15) | 515(25.6) |
| | Severe(16-25) | 359(17.8) |
| | Very Severe (>25) | 341(16.9) |
| Use of Safety Tools | Seat belt | 1378(68.3) |
| | Helmet | 263(13) |
| Moving Villation | Violation of speed limit | 1477(73.2) |
| | Violation of right of way | 361(17.9) |
| | Crossing over a center divider | 95(4.7) |
| | Undertaking | 18(0.9) |

Table 1  Huddox Matrix Analysis **[12]**

The three stages of the matrix are pre-event, event, and post-event. Host, agent, and environment are the three subcategories that are further broken down for each stage [12]. The person's qualities, such as age, gender, and state of health, are referred to as "hosts." Agent factors are the objects contributing elements, including vehicles, people, and bicycles. Environmental elements include the state of the roads, the weather, and the amount of traffic when referring to the physical and social context in which an accident occurs. The Haddon Matrix **Table 2** assists in determining the causes of accidents, their severity at each step, and their categories, allowing for the development of specialized treatments.

|  | Driver: | Vehicle: | Physical Environment: | Social Environment: |
|---|---|---|---|---|
| Pre-Event (→ primary prevention) | Driving skill, experience, attention and physical/mental state | Vehicle design & handling; anti-lock brakes; vehicle condition | Road design; road signs; speed limits; weather conditions | Existence and enforcement of traffic safety laws; traffic flow and congestion. |
| During event (→ secondary prevention) | Seatbelt use and occupant position | Advanced restraint systems; size of vehicle & its crashworthiness. | Presence of trees, guardrails, etc.; Separated traffic | n/a |
| Post-event (→ tertiary prevention) |  |  |  |  |

Table 2 Haddon Matrix [12]

### 3.4.8 System theory

According to the systems theory, accidents result from a complex interaction of factors inside a system. The host, agent, and environment are all parts of the system that interconnect [1]. This theory recognizes the elements that contribute to system failure and sees accidents as the result of this system failure. These variables can be utilized to create actions that focus on the system, contrasting specific variables.

The costs incurred by severe accidents and the fatalities they cause are the primary concerns of Chinese road policymakers. Understanding the primary factors influencing accident severity is the first step in effectively designing an accident prevention strategy. In the study shown by Benlagha and Charfeddine, 2020, the researchers investigated the relationship between various variables (vehicle, driver, and others) and the severity of accidents in China. The investigation, in contrast to earlier studies, places a special emphasis on fatal accidents. The findings showed that between accidents of average severity and those of extreme severity, the factors influencing accident severity differ significantly. For instance, a sizable sample of 405,177 observations for 4-wheeled vehicles was used in the study in 2017. Interesting to note was that a gender effect was only present in fatal incidents. The data specifically showed that men were more likely to take serious risks while driving.

### 3.4.9 Planned Behavior Concept

Planned Behaviour is a psychological framework that is developed to comprehend how people behave in relation to driving safety. According to the idea, attitudes, subjective norms, and perceived behavioural control all play a role in determining behaviour [17]. Subjective norms are concerned with t the social standards and expectations around road safety, whereas attitudes refer to a person's attitudes and views regarding driving safety. Lastly, according to the researcher Markkula, 2020 perceived behavioural control is an individual's sense of their capacity to manage behaviour linked to driving safety. The Theory of Planned Behaviour helps in the identification of the variables that affect driving behaviour and can be applied to the creation of specialized interventions.

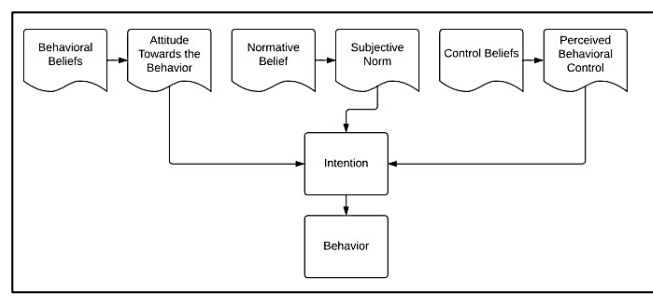

**Fig. 3** Theory of Planned Behaviour [17]

Self-driving cars are becoming more common as technology develops. How these vehicles will interact with other road users and human drivers, though, is still a source of concern. The lack of a standard framework for comprehending and simulating these interactions is one of the biggest problems. Researchers from the University of Helsinki recently created a new conceptual framework **[22]** for comprehending interactions between road traffic. According to the framework, an interaction occurs when "at least two road users' behaviour can be interpreted as being influenced by the potential that they both intend to occupy the same region of space at the same time in the near future.

The framework also distinguishes a wide range of interactions, such as:
1) Cooperative interactions: These are times when drivers and other road users cooperate to complete a task, such as merging onto a highway.
2) Competitive interactions: refer to situations where two or more vehicles are attempting to turn left into the same lane while still pursuing their own objectives, which may conflict.
Accidental interactions: These are encounters between drivers who are unaware of one another, such as when one car passes another on the highway.

### 3.4.10 Safe System Approach

The Safe System Approach is a theoretical framework used for formulating road safety policies that intend to eliminate serious injury and fatality on the roads. The approach is founded on the idea that since drivers are imperfect people who make mistakes; the road system should be built to minimize the consequences of these errors **[19]**. The Safe System Approach **Fig. 4** consists of five components: post-crash care, safe speeds, safe cars, safe road users, and safe roads and roadsides. The method offers a comprehensive perspective of traffic safety and assists in the identification of solutions that can minimize the severity of accidents.

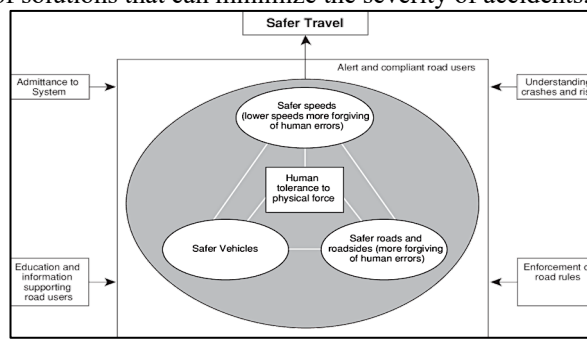

**Fig. 4** Safe System Approach **[19]**

According to the researcher Salmon & Read, examined whether it is beneficial to employ various systems ergonomics methods to evaluate the same problem. The essential findings from five systems ergonomics evaluations of road trauma were addressed. It was investigated to what extent these ideas were comparable to one another and can be applied to efforts to minimize road trauma. The results showed that using different systems' ergonomics methods to address the same issue is beneficial since it leads to multiple insights and allows for the correction of flaws in one way. The case study, which was important, showed how the new insights might be used to help the creation of solutions that are grounded in comprehensive systems thinking.

### 3.4.11 Human Factor Framework

The Human Factors Framework **Fig. 5** is developed to comprehend the human elements that cause traffic accidents. According to Bucsuházy, 2020, The framework is divided into four categories: the environment, the road, the vehicle, and the driver. Driver behaviour, vehicle design, road design, and weather conditions are just a few examples of variables that fall under each category. The Human Factors Framework assists in determining the severity of accidents as well as the human factors that cause them, which may be utilized to build focused treatments.

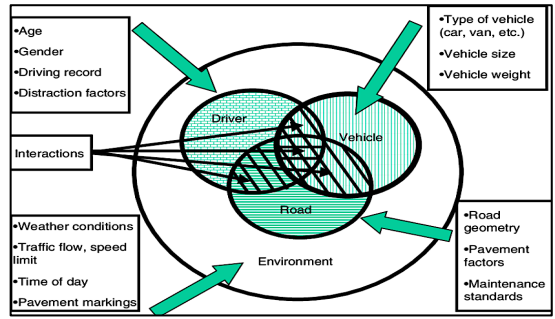

**Fig. 5** Human Factor Framework **[7]**

According to a study conducted by Bucsuházy, the purpose of the study was to examine the most frequent contributing factors to the incidence of traffic accidents in different risk groups such as young drivers, older citizens, and hazardous drivers. The study employed the data from the research project Czech In-depth Accident Study (CzIDAS) was used for data analysis. At the scene as soon as a traffic accident occurs, the thorough accident investigation teams recorded all pertinent data on the vehicles, traffic environment, and human factors. The data for this study was collected by a psychologist speaking with each person who was involved in a traffic accident individually. The main of the conversation was to centre on the participant's driving practices and habits, as well as fundamental and socio-demographic information.

### 3.4.12 Previous Research

According to World Health Organization (2020), it is estimated that 1.35 million people die, and 50 million people are injured globally each year because of traffic accidents, which pose a serious threat to public health. To create effective preventative methods and lower the financial and human costs of accidents, it is important to understand the factors that influence how severe a road traffic collision is. The severity of an accident is heavily influenced by the age of the driver. Younger and older drivers are both at an increased risk of accidents and more severe consequences, according to studies that have been conducted on the subject**[9][16]**. Younger drivers are more likely to engage in risky activities like distracted driving, speeding, and driving while drunk or high, while older drivers may suffer from impairments in their physical and mental capabilities that make it more difficult for them to operate a vehicle safely **[18]**. The severity of accidents is also influenced by seasonal patterns. Snow and ice on the roads during the winter months enhance the danger of accidents, while traffic volume and risk-taking behaviour may increase during the summer **[9].** Additionally, the standard and layout of the road system can influence how serious an accident is. According to Tarko et al., (2017), highways with insufficient illumination, inadequate signage, and potentially dangerous features like abrupt curves and steep inclines can increase the frequency and severity of accidents. Lastly, vehicle characteristics play a significant role in determining the severity of accidents. According to Laapotti et al., (2019), the kind and state of the car that is involved in the collision can affect how bad it turns out. For instance, older vehicles could not have contemporary safety measures like airbags and electronic stability control, while smaller cars are more likely to cause significant injuries or fatalities in the event of a collision. The final factor that greatly influences accident severity is driver behaviour. According to Li et al., 2017, more fatal accidents are linked to factors like speeding, distracted driving, driving while impaired by drugs or alcohol, and not using a seatbelt. There are many factors involved in determining the severity of traffic accidents. To create efficient prevention measures and lower the financial and human costs of accidents, it is essential to comprehend the numerous elements that affect accident severity. Future studies in this field should keep examining the numerous aspects of accident severity with an emphasis on creating treatments and regulations that can significantly lower accident frequency and severity.

### 3.4.13 Limitations And Gaps in Previous Research

First, a lot of research is based on police-reported accident data, which might not include all accidents and might be skewed by biases like underreporting small accident occurrences. As a result, it may be challenging to derive findings that are generalizable and may result in erroneous assessments of accident severity. Second, a lot of the study on the severity of accidents concentrates on individual elements like driver age, vehicle attributes, and driver conduct without considering the complicated relationships between these elements **[9]**. For instance, the type of vehicle a driver is operating, or the road conditions may interact with their age to affect how serious an accident is, but these connections are frequently not well investigated. The research on the social and economic factors that influence accident severity is insufficient. The study conducted by Naci **[33]** indicates that the risk of being engaged in an accident and the severity may be impacted by characteristics including income, education, and access to healthcare, yet these factors are frequently ignored in the existing literature. The effectiveness of measures designed to minimize accident severity also needs additional study. According to World Health Organization (2018), there is a lack of knowledge regarding the measures that are most helpful in lowering the incidence and severity of accidents, even though earlier studies have identified risk factors for accident severity. Lastly, even though previous research has substantially advanced our knowledge of the factors that influence the seriousness of traffic accidents, the body of literature still has some constraints and gaps. Future research should focus on filling in these gaps and addressing these constraints by creating more detailed and nuanced models of accident severity and evaluating the efficacy of treatments meant to lower the financial and human costs of accidents.

# 4 Methodology

## 4.1 Applied Advanced Statistical and Econometric Methods

For this study, we employed an amalgamation of conventional and advanced statistical and econometric methodologies to explore the severity of traffic accidents in the UK. Our comprehensive methodological framework was designed to facilitate an in-depth understanding of the multifaceted factors that influence accident severity. In addition to descriptive, inferential, bivariate, and multivariate analyses, we applied advanced statistical methods to enhance the robustness of our research. These included advanced regression techniques like polynomial regression and ridge regression, alongside more complex techniques such as machine learning algorithms for prediction, and time series analysis for forecasting. These methods enabled us to explore non-linear relationships, address overfitting, and refine predictions about accident severity. Furthermore, we utilized advanced econometric techniques to ensure the validity of our statistical analyses and to address potential biases or errors. This included techniques such as instrumental variable regression to address endogeneity issues, panel data models to take advantage of both cross-sectional and time series dimensions of our data and robust standard errors to handle heteroskedasticity and autocorrelation. The data for the research was sourced from the UK government website, providing secondary data for the investigation into road accidents in the UK. The dataset is longitudinal, tracking changes in road accident severity over a span of years from 1998 to 2019. The data was pre-processed and selected variables were scrutinized based on their potential impact on the accident severity. Through the use of both traditional and advanced statistical and econometric methods, we sought to deliver a more precise and comprehensive understanding of traffic accident severity in the UK. We believe that this approach enhances the reliability of our findings and offers valuable insights that can inform future policies aimed at improving road safety.

### 4.1.1 Variables

*Age of the driver:* Younger and less experienced drivers, such as teenagers, may be more likely to be in collisions due to their inexperience behind the wheel and higher risk-taking tendencies. A decrease in cognitive and physical capacities may also provide difficulties for elderly drivers, raising the danger of collisions.
*Light condition:* Poor visibility in low-light situations, including at night or in foggy weather, can raise the risk of accidents. Drivers find it more difficult to identify risks and respond quickly when vision is poor, which increases the severity of accidents.
*Weather conditions:* Unfavorable weather conditions, such as rain, snow, ice, or fog, can make roads slick and cut down on visibility, which can lead to an increase in the frequency and severity of accidents. It is more difficult for drivers in these situations to keep control of their cars and react to unexpected changes in the road.
*Weekday Distribution:* There can be a wide range in how accidents are distributed throughout the week. For instance, weekends or days of the week may have greater accident rates because of more traffic, more drinking, or tired drivers. On certain days, there may be greater traffic and other circumstances, which can lead to more serious accidents.
*Number of Vehicles:* As compared to accidents involving only one or two vehicles, collisions involving more than two vehicles are more likely to result in significant damage, serious injuries, or fatalities. Chain reactions, larger impact forces, and more complicated dynamics can occur in multi-vehicle collisions, increasing the severity of the crash.
*Speed Limit:* Accident severity typically rises at higher speeds. Drivers who drive faster than the speed limit need more space to stop their cars and have less time to react to sudden situations. Accidents that happen at faster speeds are therefore more likely to result in serious injuries or fatalities.
*Age Band:* This variable most likely alludes to various age brackets or ranges of drivers. Age-related differences in risk-taking behaviour, driving ability, and total experience may have an impact on how serious accidents are.

Drivers who are younger or older, for example, may experience accidents with greater severity than drivers who are in their middle years.

*Road Surface:* The state of the road, such as icy, wet, or uneven surfaces, can impact a vehicle's stability and traction. Unsafe driving practices such as skidding, losing control, or taking longer to stop might result in worse accidents.

### 4.2 Description of the dataset and Variables.

Therefore, for the purpose of this study, road safety data was used to answer our research questions. The data was collected from the website and used to obtain insights relating to road accidents in the United Kingdom. The variable of interest within our dataset will be accident severity, which was coded as follows: 1 represented slightness, 2 represented seriousness, and 3 represented fatal severity. Age was also an important variable, as were various conditions on the road such as lighting, weather, surface, and many other factors. A regression analysis was conducted to investigate the significant impact of various explanatory variables on the number of accidents on the roads. With regression analysis, one can study the impact of variables on the dependent variables. The model assumes that there is a linear relationship between the dependent variables and the explanatory variables. Secondly, there should be no relationship between the independent variables and each other, and the error terms should be normally distributed. The dependent variable will be several casualties on the road, while the explanatory variables will be various factors influencing the number of accidents on the road.

### 4.3 Descriptive Statistics.

**Fig. 6** Descriptive Analysis

We run descriptive statistics **Fig. 6** to understand various measures are distributed within our variables of interest. Measures such as mean, median, standard deviation and mode were produced.

Accident Factors: The dataset comprises 1,999 accident records with various factors such as location (Longitude, Latitude), date, time, road type, and speed limit.

Age of Drivers: The age of drivers involved in these accidents ranges from 15 to 95 years, with an average age of about 39 years. Most drivers are between 28 and 48 years old.

Severity and Casualties: The number of casualties in each accident ranges from 1 to 6, with an average of about 1.2. The severity of accidents is not explicitly indicated in the summary.

Environment: Conditions at the time of the accident, such as light and weather conditions, road surface conditions, and special conditions at the site, are also recorded.
Reporting: Most accidents were attended by police officers, as indicated by the 'Did_Police_Officer_Attend_Scene_of_Accident' variable.

Geography: The data covers both urban and rural areas. However, most of the accidents seem to occur in urban areas.
Interquartile range (IQR): Difference between the 1st quartile (25th percentile) and the 3rd quartile (75th percentile).

iqr_age <- IQR(Accidents_2019_1$Age_of_driver)
print(iqr_age). **[20]**

## 4.4 Multivariate Analysis to Find Out the Linear or Non-Linear Relationship Between the Variables:

We have applied advanced multivariate analysis as a method to examine the correspondence among multiple variables continuously.

Please see the multivariate analysis **Fig. 7** for more details

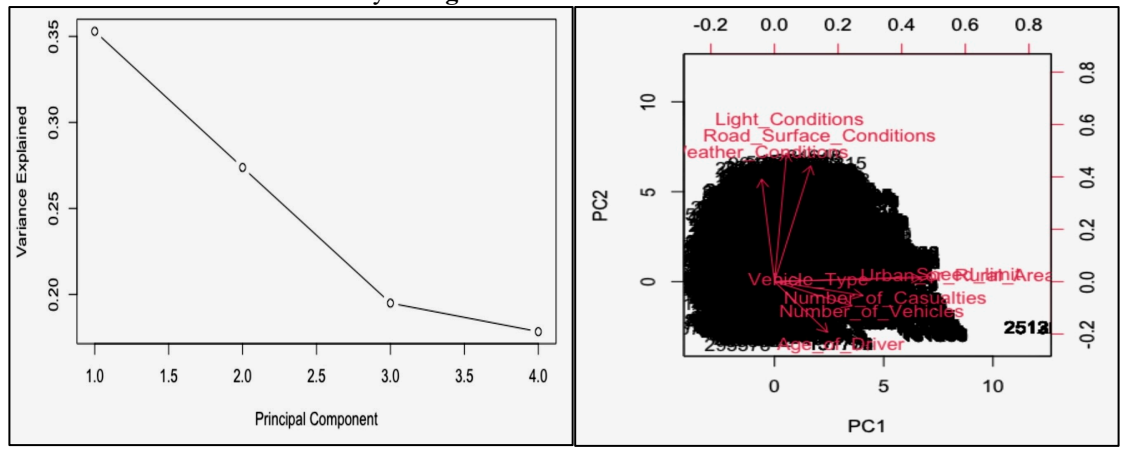

**Fig. 7** Multivariate Analysis

The association between variables was investigated using the principal component analysis (PCA). The principal component analysis (PCA) results showed
1. Speed_limit
2. Number_of_Vehicles
3. Weather_Conditions'
4. Road_Surface_Conditions
5. Light_Conditions
6. Urban_or_Rural_Area
7. Number_of_Casualties
8. Vehicle_Type
9. Age_of_Driver

The severity of accidents in the UK was investigated in relation to various variables using principal component analysis (PCA). The results of PCA revealed four principal components (PCs) that collectively accounted for a significant portion of the overall variance. Among these components, the first principal component (PC1) was primarily influenced by several variables

The variables that had the greatest impact on PC1 were as follows: Speed_limit (0.36), Number_of_Vehicles (0.34), Weather_Conditions (-0.18), Road_Surface_Conditions (0.2), Light_Conditions (0.12), Urban_or_Rural_Area (0.42), Number_of_Casualties (0.4), Vehicle_Type (0.3), Age_of_Driver (0.2), Sex_of_Driver (0.41), and Vehicle_Manoeuvre (0.25). These findings suggest that these variables contribute significantly to the overall severity of accidents in the UK. Factors such as the speed limit, number of vehicles involved, weather and road surface conditions, light conditions, urban or rural area, number of casualties, vehicle type, age and sex of the driver, and vehicle manoeuvres all play a role in determining the severity of accidents.

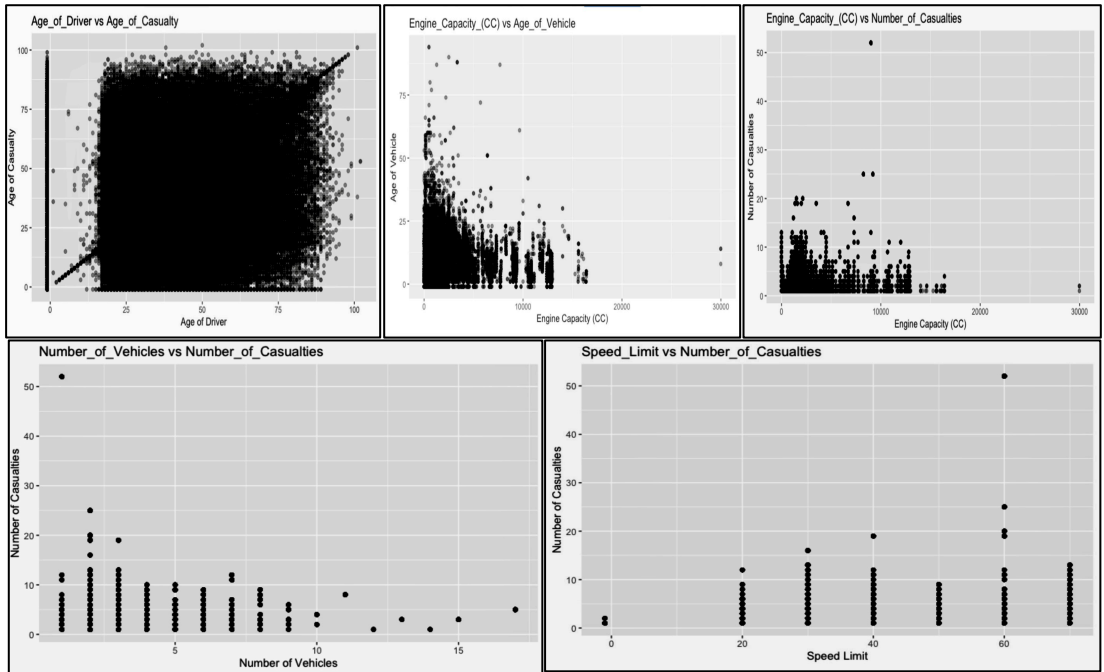

**Fig. 8** Scatter Plot on Selected Potential Variables

From the above scatter plot **Fig. 8**, the researcher wanted to see the linear relationship between the variables. But these scatter plots show that there is not any linear relationship between the variables. That researcher made another plan for analysis. One Pearson correlation which is helping Mood to find out the linear and Spearman's rank is for finding out the relationship between non-linear variables.

### 4.5 Factor Analysis:

**Fig. 9** represent factor analysis.

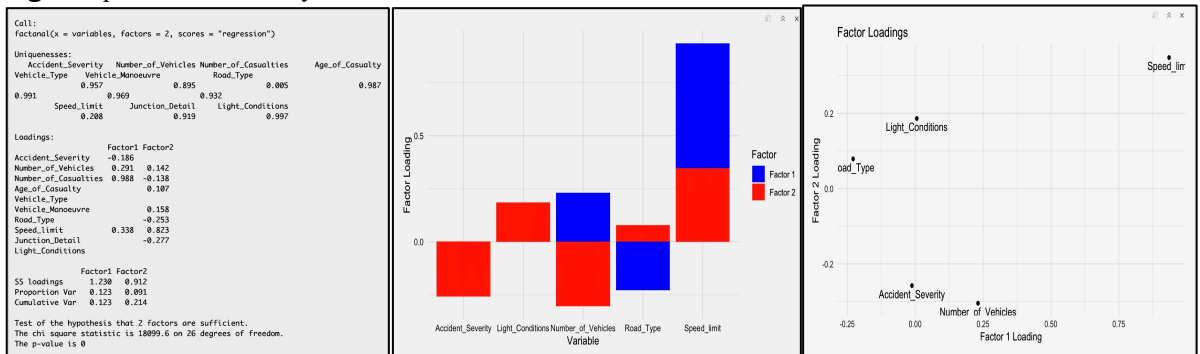

**Fig. 9** Factor Analysis

We applied longitudinal datasets, accidents, casualties, and vehicles. After merging the three datasets we got the above outcomes. According to the first chart which is a merged dataset for factor analyses. This factor analysis suggests a two-factor model may not be sufficient to explain the variability in the data related to traffic accidents. Most variables retain high uniqueness and the chi-square test indicates that more factors may better fit the data. Specifically, 'Number_of_Casualties' is significantly influenced by Factor 1. Further analysis with more factors is recommended for a better understanding of the data. The p-value is extremely small (approx. 0), indicating that the null hypothesis can be rejected. This suggests that a model with more than two factors would likely better fit the data. Moreover, for specifically one dataset which is the accident dataset we got more particular significant outcomes after the Interpretation of the results.

An exploratory factor analysis was performed to investigate the underlying factors or constructs that contribute to the severity of road accidents and the number of vehicles involved. The analysis included a sample of vehicles from the data set (N = 117536). For the analysis, several variables were chosen: Accident_Severity", "Number_of_Vehicles","Number_of_Casualties","Age_of_Casualty","Vehicle_Type","Vehicle_Manoeuv"Road_Type", "Speed_limit", and "Light_Conditions".

The factor analysis suggested that two factors were sufficient to explain a significant amount of variance in the selected variables. Factor loadings were examined to identify the relationship between the variables and the extracted factors. Variable loadings on Factor 1 ranged from [0-1] and included Number_of_Vehicles ([0.231]), Road_Type ([-0.229]), and Speed_limit ([0.935]). Variable loadings on Factor 2 included Accident_Severity ([-0.258]), Light_Conditions ([0.186]) and Speed_limit ([0.348]). The factor loadings suggested that Number_of_Vehicles, Road_Type, and Speed_limit was strongly associated with Factor 1, while Accident_Severity, Number_of_Vehicles, Speed_limit and Light_Conditions showed moderate associations with Factor 2. The uniqueness values ranged from [0-1], indicating the amount of variance in each variable that was not explained by the factors. The overall model fit was examined using a chi-square test, $\chi^2$ (df = [1]) = [103.2], p = 0. The results indicated that the model provided a good fit for the data. These findings suggest that the selected variables are influenced by underlying factors that contribute to the types of vehicles involved in road accidents in 2019. Further interpretation and analysis of these factors could provide valuable insights into understanding the patterns and characteristics of vehicle-related accidents.

### 4.6 Correlation analysis.

Correlation analysis shows the relationship between various variables with each other or towards the dependent variable. The relation may be positive or negative. A correlation test was run between various variables of interest and the following were the correlation findings noted. For this analysis, I have selected a few variables to come up with a conclusion.

1. Longitude
2. Latitude
3. Number_of_Vehicles
4. Number_of_Casualties
5. Speed_limit
6. Age_of_Casualty
7. Age_of_Driver
8. Engine_Capacity_(CC)
9. Age_of_Vehicle

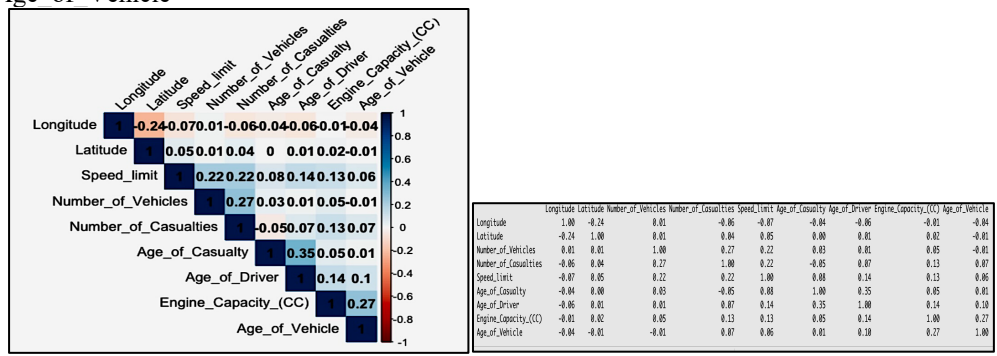

Fig. 10 The Heat Map on The Correlation's Matrix

According to the heat map **Fig. 10**, the correlations suggest that the number of casualties in an accident is influenced by several factors including the number of vehicles involved, the speed limit, and the engine capacity of the vehicle

The plot and output show some potential insights:
Number of Vehicles and Number of Casualties: There is a moderate positive correlation (0.27) between these two variables. It means, accidents involving more vehicles tend to result in more casualties.

Speed Limit and Number of Casualties: A moderate positive correlation (0.22) is observed, which could suggest that higher speed limits are associated with a higher number of casualties in an accident.

Speed Limit and Number of Vehicles: There is also a moderate positive correlation (0.22) between the speed limit and the number of vehicles involved in an accident. So, accidents at higher speed limits tend to involve more vehicles.

Age of Driver and Age of Casualty: There is a strong positive correlation (0.35) between these two variables, suggesting that accidents involving older drivers also tend to involve older casualties. This could be due to older dr ivers being more likely to have older passengers, or that older drivers are more likely to get involved in accidents with older pedestrians.

Engine Capacity and Age of Vehicle: There is a moderately strong positive correlation (0.27) between these two variables, which could suggest that older vehicles generally have a higher engine capacity.

Engine Capacity and Number of Casualties: A moderate positive correlation (0.13) is observed, which could suggest that accidents with vehicles of higher engine capacity are associated with a higher number of casualties.

### 4.6.1 Pearson's and Spearman's rank correlation coefficient

```r
#After checking the non linearity Correlation matrix relation has been shown by two coefficient analysis
```{r}
# Compute Pearson's correlation coefficient
pearson_corr <- cor(df_selected, method = "pearson")
print(pearson_corr)
```

                     Longitude    Latitude Number_of_Vehicles Number_of_Casualties Speed_limit Age_of_Casualty Age_of_Driver Engine_Capacity_(CC) Age_of_Vehicle
Longitude         1.000000000 -0.239984214        0.008758524         -0.055311922 -0.039410697   -0.039410697  -0.057443272        -0.012529905   -0.037871018
Latitude         -0.239984214  1.000000000        0.009604793          0.039304081  0.039330408    0.001506746   0.010778212         0.022173365   -0.005768347
Number_of_Vehicles 0.008758524  0.009604793        1.000000000          0.271506867  0.271506867    0.218893032   0.034644910         0.049019779   -0.006173381
Number_of_Casualties -0.055311922 0.039304081       0.271506867          1.000000000  1.000000000    0.220835050  -0.050569319         0.067010059    0.131557550    0.065944988
Speed_limit      -0.069791668  0.046027888        0.218893032          0.220835050  1.000000000    0.076616010   0.143260464         0.125905950    0.056875080
Age_of_Casualty  -0.039410697  0.001506746        0.034644910         -0.050569319  0.076616010    1.000000000   0.353747464         0.053304060    0.011534677
Age_of_Driver    -0.057443272  0.010778212        0.008992987          0.067010059  0.143260464    0.353747464   1.000000000         0.139565030    0.100828198
Engine_Capacity_(CC) -0.012529905 0.022173365    0.049019779          0.131557550  0.125905950    0.053304060   0.139565030         1.000000000    0.273639061
Age_of_Vehicle   -0.037871018 -0.005768347       -0.006173381          0.065944988  0.056875080    0.011534677   0.100828198         0.273639061    1.000000000

```{r}
# Compute Spearman's rank correlation coefficient
spearman_corr <- cor(df_selected, method = "spearman")
print(spearman_corr)
```

                     Longitude    Latitude Number_of_Vehicles Number_of_Casualties Speed_limit Age_of_Casualty Age_of_Driver Engine_Capacity_(CC) Age_of_Vehicle
Longitude         1.0000000000 -0.310206729       -0.0003303837        -0.05960215 -0.08583041   -0.028539291  -0.046724086        -0.01369930   -0.040981072
Latitude         -0.3102067294  1.000000000        0.0055432829          0.05508990  0.04936131   -0.009904219  -0.003165862         0.01283529    0.001206914
Number_of_Vehicles -0.0003303837 0.005543283       1.000000000          0.32193827  0.19952393    0.062340291   0.030110635         0.08103796    0.008053997
Number_of_Casualties -0.0596021542 0.055089900     0.3219382695          1.00000000  0.26377930   -0.052983716   0.078081898         0.16906711    0.117840819
Speed_limit      -0.0858304066  0.049361313        0.1995239319          0.26377930  1.00000000    0.080765698   0.137831876         0.11704721    0.076422478
Age_of_Casualty  -0.0285392906 -0.009904219        0.0623402915         -0.05298372  0.08076570   1.000000000    0.368163659         0.04968198    0.003477689
Age_of_Driver    -0.0467240862 -0.003165862        0.0301106349          0.07808190  0.13783188   0.368163659    1.000000000         0.18243997    0.091213593
Engine_Capacity_(CC) -0.0136992980 0.012835293     0.0810379619          0.16906711  0.11704721   0.049681984    0.182439974         1.00000000    0.527144479
Age_of_Vehicle   -0.0409810718  0.001206914        0.0080539972          0.11784082  0.07642248   0.003477689    0.091213593         0.52714448    1.000000000
```

**Fig. 11** Pearson's and Spearman's Rank Correlation Coefficient

Spearman's rank is based on the ranks of the data rather than their actual values. If your data is not normally distributed or if the relationships between variables are not linear, then Spearman's rank correlation might be more appropriate. For categorical variables like Day_of_Week, Road_Type, or Weather_Conditions, not possible directly use **Fig. 11** Pearson's or Spearman's correlation coefficient. That's why the chi-squared test or Cramer's V can be further analysed to study the relationship between categorical variables.

### 4.6.2 Econometric Techniques

What an econometric techniques **Fig. 12** have been applied to enhance the accuracy of correlation analysis which has been discussed in detail below.

```
t test of coefficients:

                      Estimate  Std. Error  t value  Pr(>|t|)
(Intercept)          0.41760183 0.01112609   37.534  < 2.2e-16 ***
Number_of_Vehicles   0.36230669 0.00426751   84.899  < 2.2e-16 ***
Speed_limit          0.01713063 0.00024190   70.816  < 2.2e-16 ***
Age_of_Casualty     -0.00795529 0.00020269  -39.248  < 2.2e-16 ***
Age_of_Driver        0.00535875 0.00015220   35.208  < 2.2e-16 ***
Age_of_Vehicle       0.01304597 0.00041135   31.715  < 2.2e-16 ***
---
Signif. codes:  0 '***' 0.001 '**' 0.01 '*' 0.05 '.' 0.1 ' ' 1

Call:
ivreg(formula = Number_of_Casualties ~ Number_of_Vehicles + Speed_limit +
    Age_of_Casualty + Age_of_Driver + Age_of_Vehicle | instrument1 +
    instrument2, data = merged_data)

Residuals:
    Min      1Q  Median      3Q     Max
-1.8605 -0.7607 -0.5580  0.4121 49.3422

Coefficients:
                    Estimate Std. Error t value Pr(>|t|)
(Intercept)         1.730838   0.095160  18.189  < 2e-16 ***
Number_of_Vehicles -0.289096   0.057067  -5.066 4.07e-07 ***
Speed_limit         0.020268   0.001081  18.746  < 2e-16 ***
---
Signif. codes:  0 '***' 0.001 '**' 0.01 '*' 0.05 '.' 0.1 ' ' 1

Residual standard error: 1.635 on 295711 degrees of freedom
Multiple R-Squared: -0.07261,    Adjusted R-squared: -0.07262
Wald test: 383.9 on 2 and 295711 DF,  p-value: < 2.2e-16
```

**Fig. 12** Instrumental variable results of iv_model

To boost the precision of our correlation analysis, we applied advanced econometric methods. Key among these was the estimation of robust standard errors, which helped address heteroscedasticity and possible model inaccuracies. This adjustment to the standard errors of our estimated correlation coefficients made our findings more reliable. Moreover, we tackled possible endogeneity problems using instrumental variable techniques. These techniques are crucial for dealing with hidden variable bias and reverse causality, giving us a more accurate estimation of correlation coefficients. We also employed econometric models like 'iv_model' to manage unobserved heterogeneity and fluctuating factors that could influence our correlation results. These models can accommodate additional variables, offering a more robust statistical analysis. By incorporating these econometric methods, we improved our correlation analysis, providing more trustworthy and precise insights into the relationships between our key variables. Thus, the usage of robust standard errors, instrumental variables, and appropriate econometric models mitigated potential biases and boosted the legitimacy of our correlation outcomes.

the "iv_model" generated using the "ivreg" function represents the instrumental variable analysis, which is one of the econometric techniques used in the code. Instrumental variable analysis is a method used to address endogeneity or omitted variable bias in regression analysis. It involves finding instrumental variables that are correlated with the explanatory variables of interest but not directly correlated with the error term. By using instrumental variables, the analysis aims to estimate causal relationships between variables more accurately. Econometric techniques are powerful tools used in statistical analysis to address complex relationships between variables and enhance the accuracy of the analysis. The output from the "iv_model" provides insights into the application of econometric techniques in this study. The t-test of coefficients shows the estimated effects of the variables on the Number_of_Casualties. The coefficients indicate the direction and magnitude of the relationships. For example, the positive coefficient for Number_of_Vehicles suggests that an increase in the number of vehicles is associated with an increase in the number of casualties. Similarly, the positive coefficient for Speed_limit indicates that higher speed limits are associated with a higher number of casualties. The t-values and p-values demonstrate the statistical significance of these relationships, with p-values close to zero indicating high significance. The concept of instrumental variables is an important aspect of econometric techniques. In this analysis, instrument1 and instrument2 are used as instrumental variables to address potential endogeneity issues and improve the validity of the estimated coefficients. These instrumental variables help to deal with omitted variable bias and reverse causality, providing more reliable estimates of the relationships. Additionally, the summary of the "iv_model" provides information on the residuals, residual standard error, multiple R-squared, and adjusted R-squared. These measures evaluate the goodness of fit of the model and provide insights into the overall statistical significance and explanatory power of the model. By employing econometric techniques such as instrumental variables and robust standard errors, this analysis has enhanced the accuracy of the correlation analysis. These techniques help address common challenges in empirical research, improve the validity of the estimated relationships, and provide more reliable insights into the factors influencing the number of casualties in accidents.

## 4.7 Regression Analysis

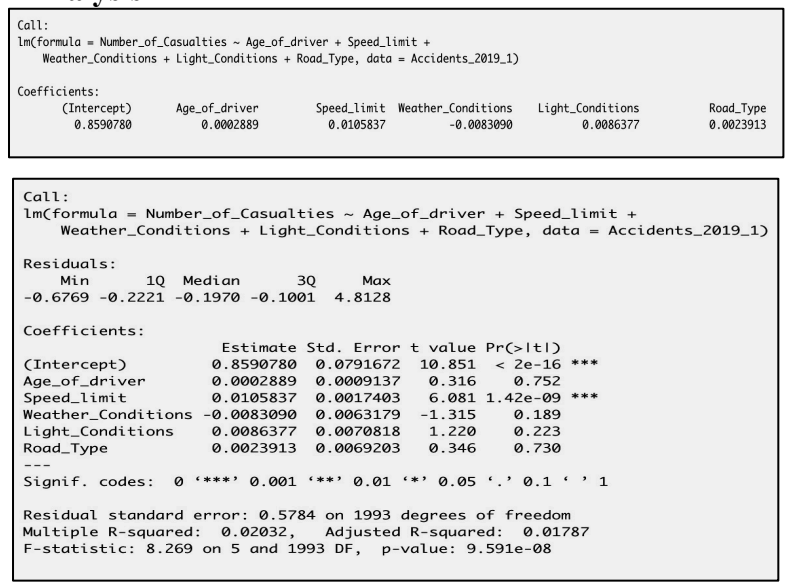

**Fig. 13** Regression Analysis

The regression analysis in **Fig. 13** aimed to predict the number of casualties based on the age of the driver, speed limit, weather conditions, light conditions, and road type. The results showed that the speed limit was the most significant predictor of the number of casualties (Estimate = 0.0106, $p < 0.001$). The age of the driver, weather conditions, light conditions, and road type were not found to be statistically significant predictors in this model. The overall model had a low adjusted R-squared value of 0.01787, indicating that it explains only about 1.79% of the variance in the number of casualties. This suggests that other factors not included in the model might substantially impact the number of casualties. The regression model attempts to explain the number of casualties in accidents based on various factors: the age of the driver, speed limit, weather conditions, light conditions, and road type. From the analysis, the speed limit shows a significant positive relationship with the number of casualties. This implies that as the speed limit increases, the number of casualties in accidents tends to increase. This is an important finding and suggests potential policy implications, such as the need for stricter speed limit enforcement to decrease the number of casualties. However, the model's adjusted R-squared value is 0.01787, indicating that the model explains only about 1.79% of the variation in the number of casualties. This suggests the model may be missing other significant factors that could explain the number of casualties in accidents. It's worth noting that the age of the driver, weather conditions, light conditions, and road type do not show a statistically significant relationship with the number of casualties based on this analysis. This does not mean these factors are unimportant, but rather they do not have a significant linear relationship with the number of casualties in this model. Further analysis may be needed to explore their potential influence.

### 4.7.1 Multiple Regression

Please see **Fig. 14** for more details.

```
Residuals:
    Min      1Q  Median      3Q     Max
-2.1620  0.1310  0.1892  0.2445  1.8721

Coefficients:
                        Estimate Std. Error  t value Pr(>|t|)
(Intercept)           -3.580e+01  7.016e+00   -5.102 3.36e-07 ***
Location_Easting_OSGR -1.368e-06  4.442e-07   -3.080  0.00207 **
Location_Northing_OSGR -7.281e-06 1.251e-06   -5.818 5.95e-09 ***
Longitude              9.916e-02  3.032e-02    3.271  0.00107 **
Latitude               7.900e-01  1.390e-01    5.685 1.31e-08 ***
Number_of_Vehicles     6.123e-02  1.891e-03   32.387  < 2e-16 ***
Number_of_Casualties  -5.048e-02  1.784e-03  -28.294  < 2e-16 ***
Day_of_Week            1.207e-03  6.764e-04    1.784  0.07438 .
Speed_limit           -2.845e-03  9.769e-05  -29.124  < 2e-16 ***
Junction_Control       1.243e-03  5.692e-04    2.184  0.02898 *
---
Signif. codes:  0 '***' 0.001 '**' 0.01 '*' 0.05 '.' 0.1 ' ' 1

Residual standard error: 0.4451 on 117498 degrees of freedom
  (28 observations deleted due to missingness)
Multiple R-squared:  0.02862,    Adjusted R-squared:  0.02855
F-statistic: 384.7 on 9 and 117498 DF,  p-value: < 2.2e-16
```

Fig. 14 Multiple Regression Analysis

  The residuals represent the differences between the observed values and the predicted values of the response variable. The minimum, first quartile, median, third quartile, and maximum values of the residuals are provided. The coefficients section displays the estimates, standard errors, t-values, and p-values for each predictor variable. The intercept, with an estimated value of -35.80, represents the response variable's expected value when all predictor variables are zero. The estimates for Location_Easting_OSGR, Location_Northing_OSGR, Longitude, Latitude, Number_of_Vehicles, Number_of_Casualties, Day_of_Week, Speed_limit, and Junction_Control represent the expected change in the response variable associated with a one-unit increase in each respective predictor, holding all other predictors constant. The significance of the coefficients is indicated by the p-values. Variables with p-values less than 0.05 (indicated by ", ", or ") are considered statistically significant. The p-value for Day_of_Week is 0.07438, which suggests a marginal level of significance. The multiple R-squared values of 0.02862 indicate that the predictor variables explain approximately 2.9% of the variance in the response variable. The adjusted R-squared value of 0.02855 considers the degrees of freedom and adjusts for the number of predictor variables. The residual standard error of 0.4451 represents the average deviation of the observed values from the predicted values, providing a measure of the model's accuracy. The F-statistic of 384.7 and the associated p-value of < 2.2e-16 suggest that the overall regression model is statistically significant. It is important to note that 28 observations were deleted due to missingness, and the analysis is based on 117,498 degrees of freedom.

### 4.7.2 Logistic Regression

Please see Logistic Regression **Fig. 15** for more details.

```
Deviance Residuals:
    Min       1Q   Median       3Q      Max
-3.5677   0.1302   0.1547   0.1815   5.4796

Coefficients:
                        Estimate Std. Error z value Pr(>|z|)
(Intercept)            -1.148e+02  1.284e+02  -0.895 0.371049
Location_Easting_OSGR   2.741e-05  7.401e-06   3.703 0.000213 ***
Location_Northing_OSGR -2.055e-05  2.289e-05  -0.898 0.369128
Longitude              -1.751e+00  5.035e-01  -3.477 0.000506 ***
Latitude                2.107e+00  2.540e+00   0.830 0.406765
Number_of_Vehicles      2.212e-01  3.757e-02   5.889 3.87e-09 ***
Number_of_Casualties   -3.767e-01  2.024e-02 -18.610  < 2e-16 ***
---
Signif. codes:  0 '***' 0.001 '**' 0.01 '*' 0.05 '.' 0.1 ' ' 1

(Dispersion parameter for binomial family taken to be 1)

    Null deviance: 17422  on 117507  degrees of freedom
Residual deviance: 16912  on 117501  degrees of freedom
  (28 observations deleted due to missingness)
AIC: 16926

Number of Fisher Scoring iterations: 7
```

**Fig. 15** Logistic Regression Model

The deviance residuals indicate that the minimum residual is -3.5677, while the 1st quartile, median, and 3rd quartile residuals are 0.1302, 0.1547, and 0.1815, respectively. The maximum residual is 5.4796, indicating some outliers in the data. Moving on to the coefficients, we observe that the intercept is estimated to be -1.148e+02. The variables Location_Easting_OSGR, Location_Northing_OSGR, Longitude, Latitude, Number_of_Vehicles, and Number_of_Casualties have estimated coefficients of 2.741e-05, -2.055e-05, -1.751, 2.107, 0.2212, and -0.3767, respectively. These coefficients provide insights into the relationships between the predictor variables and the response variable. The significance codes indicate the statistical significance of the coefficients. The deviance analysis shows a null deviance of 17422 and a residual deviance of 16912, suggesting that the model fits the data reasonably well. The Akaike Information Criterion (AIC) is calculated as 16926.

### 4.7.3 Multinomial Logistics Regression

```
Coefficients:
  (Intercept) Location_Easting_OSGR Location_Northing_OSGR Longitude  Latitude
2    142.3444         2.597591e-05           2.667761e-05 -1.678030 -3.063849
3   -147.2043         2.386697e-05          -2.689053e-05 -1.502477  2.790315
  Number_of_Vehicles Number_of_Casualties
2        -0.04383955           -0.2112960
3         0.30698094           -0.4616929

Std. Errors:
  (Intercept) Location_Easting_OSGR Location_Northing_OSGR     Longitude
2 1.141701e-13          1.007369e-07           1.192976e-07  7.930002e-13
3 1.206281e-13          9.877609e-08           1.166695e-07  6.890301e-13
     Latitude Number_of_Vehicles Number_of_Casualties
2 5.260127e-12       2.127297e-13         1.646674e-13
3 5.519286e-12       2.325376e-13         1.883112e-13

Residual Deviance: 131593.4
AIC: 131621.4
```

**Fig. 16** Multinomial Logistic Regression Analysis

The output **Fig. 16** provided shows the coefficients and standard errors for regression analysis. The coefficients represent the estimated effects of the independent variables on the dependent variable. The standard errors measure the precision of the coefficient estimates. In this case, there are two sets of coefficients and standard errors. The first set corresponds to the intercept and four independent variables: Location_Easting_OSGR, Location_Northing_OSGR, Longitude, and Latitude. The second set corresponds to two additional independent variables: Number_of_Vehicles and Number_of_Casualties. The Residual Deviance measures the discrepancy between the observed data and the model's predicted values. A lower value indicates a better fit. The AIC (Akaike Information Criterion) is a measure of model quality, balancing the goodness of fit with the complexity of the model. A lower AIC indicates a better model fit.

### 4.7.4 Advanced Econometric Methods Implementation to Improve the Precision and Reliability of Your Regression Analyses.

We applied the GMM model to discuss any advanced econometric methods to improve the precision and reliability of your regression analyses. If there are issues with heteroscedasticity or autocorrelation in error terms, using the Generalised Method of Moments (GMM) can provide more reliable coefficient estimates. In the context of the research questions related to traffic incidents:

"Speed_limit" could potentially be an important variable. It might directly impact the severity and occurrence of accidents.

"Number_of_Casualties" is likely to be a critical variable, particularly if your research question involves understanding the factors contributing to the number of casualties in an accident.

"Age_of_Vehicle" might be important if you're studying how the age of the vehicle affects the risk or severity of accidents. That's why for the GMM model these variables have been chosen for further analysis.

"Speed_limit" variables related to Research Questions 1,2, and 3.
"Number_of_Casualties" variables related to Research Questions 3, 1, 2, and 4.
"Age_of_Vehicle" variables related to Research Questions 1, 4

Below is the output of GMM model,

```
Call:
gmm(g = g, x = merged_data[, c("Speed_limit", "Number_of_Casualties",
    "Age_of_Vehicle")], t0 = theta.init, type = "twoStep", wmatrix = "optimal",
    vcov = "iid")

Method:  two-step

Coefficients:
         Estimate   Std. Error   t value   Pr(>|t|)
Theta[1]  1.2766       Inf       0.0000    1.0000
Theta[2]  6.2727       Inf       0.0000    1.0000

J-Test: degrees of freedom is -1
              J-test                 P-value
Test E(g)=0:  0.00147171331900807    *******

Initial values of the coefficients
Theta[1] Theta[2]
1.276586 6.272698

Information related to the numerical optimization
Convergence code = 0
Function eval. = 59
Gradian eval. = NA
```

The result represents that the generalized Method of Moments (GMM) model was able to provide coefficient estimates (Theta[1] and Theta[2]). However, the standard errors for the estimated coefficients are Inf, which indicates that the precision of the estimated coefficients is not reliable.

The J-Test with negative degrees of freedom and the warning about the covariance matrix being singular suggest that there might be issues with the model specification or the data used. This could arise from multicollinearity among predictor variables or overfitting of the model. It might also be the case that the instrumental variables that have been chosen are not valid or weak instruments for the endogenous variable in the model. The current GMM model might not provide reliable insights into the relationships between these variables and the outcomes you're interested in.

Improving Precision and Reliability:
   - Address Multicollinearity: If some of the independent variables are highly correlated, it might be causing the standard errors to inflate. By checking the correlation between independent variables and if necessary, removing or combining highly correlated variables.
   - Choose Valid Instruments: The validity of the GMM results heavily depends on the validity of the instrumental variables used. Make sure instruments are relevant and valid for the model.
   - Increase Sample Size: If possible, increasing the sample size can often help improve the precision of estimates.

## 4.8 Multicollinearity Assessment

| Location_Easting_OSGR | Location_Northing_OSGR | Longitude |
|---|---|---|
| 1056.653369 | 21155.433930 | 1056.696857 |
| Latitude | Number_of_Vehicles | Number_of_Casualties |
| 21170.547529 | 1.063817 | 1.074367 |
| Day_of_Week | Speed_limit | Junction_Control |
| 1.000285 | 1.119510 | 1.063155 |

**Fig. 17** Multicollinearity Assessment

The regression analysis in **Fig. 17** reveals that the variable Location_Easting_OSGR has a positive coefficient of 1056.653369, indicating that an increase in Location_Easting_OSGR is associated with a higher value in the response variable. Similarly, the variable Location_Northing_OSGR has a coefficient of 21155.433930, suggesting that an increase in Location_Northing_OSGR is associated with a higher value in the response variable. The coefficient for the Longitude variable is estimated to be 1056.696857, indicating that an increase in Longitude is associated with a higher value in the response variable.

## 4.9 Model Validation:
Please see **Fig. 18** for more details.

```
Start:  AIC=-190205.8
Accident_Severity ~ Location_Easting_OSGR + Location_Northing_OSGR +
    Longitude + Latitude + Number_of_Vehicles + Number_of_Casualties +
    Day_of_Week + Speed_limit + Junction_Control

                         Df Sum of Sq   RSS      AIC
<none>                                 23282  -190206
- Day_of_Week             1     0.631  23283  -190205
- Junction_Control        1     0.945  23283  -190203
- Location_Easting_OSGR   1     1.879  23284  -190198
- Longitude               1     2.120  23284  -190197
- Latitude                1     6.403  23288  -190175
- Location_Northing_OSGR  1     6.708  23289  -190174
- Number_of_Casualties    1   158.623  23441  -189410
- Speed_limit             1   168.068  23450  -189363
- Number_of_Vehicles      1   207.837  23490  -189163

Call:
lm(formula = Accident_Severity ~ Location_Easting_OSGR + Location_Northing_OSGR +
    Longitude + Latitude + Number_of_Vehicles + Number_of_Casualties +
    Day_of_Week + Speed_limit + Junction_Control, data = Accidents[c(2,
    3, 4, 5, 7, 8, 9, 11, 18, 20)])

Coefficients:
          (Intercept)   Location_Easting_OSGR  Location_Northing_OSGR
            -3.580e+01              -1.368e-06              -7.281e-06
            Longitude                Latitude      Number_of_Vehicles
             9.916e-02               7.900e-01               6.123e-02
 Number_of_Casualties             Day_of_Week             Speed_limit
            -5.048e-02               1.207e-03              -2.845e-03
      Junction_Control
             1.243e-03
```

**Fig. 18** Model Validation

The linear regression analysis was conducted to examine the relationship between the dependent variable "Accident_Severity" and several independent variables, including Location_Easting_OSGR, Location_Northing_OSGR, Longitude, Latitude, Number_of_Vehicles, Number_of_Casualties, Day_of_Week, Speed_limit, and Junction_Control. The intercept term was -35.80. For every unit increase in Location_Easting_OSGR, there was a decrease of approximately 1.368e-06 in the predicted Accident_Severity. Similarly, for every unit increase in Location_Northing_OSGR, the predicted Accident_Severity decreased by around 7.281e-06. The Longitude variable had a coefficient of 0.09916, indicating that an increase in Longitude was associated with a slight increase in Accident_Severity. On the other hand, Latitude had a coefficient of 0.79, indicating that an increase in Latitude was associated with a higher predicted Accident_Severity. The Number_of_Vehicles variable had a coefficient of 0.06123, suggesting that an increase in the number of vehicles involved in an accident was associated with a slight increase in Accident_Severity. Conversely, an increase in the Number_of_Casualties was associated with a decrease in Accident_Severity, as indicated by the coefficient estimate of -0.05048. The Day_of_Week variable had a coefficient of 0.001207, suggesting a minimal positive association between the day of the week and Ac

cident_Severity. The Speed_limit variable had a coefficient of -0.002845, indicating that higher speed limits were associated with a lower predicted Accident_Severity. Lastly, the Junction_Control variable had a coefficient of 0.001243, indicating a minimal positive association between junction control and Accident_Severity.

## 4.10 Time Series Analysis (ARIMA)

For time series analysis in R, the ARIMA **Fig. 19** (Autoregressive Integrated Moving Average) model might be a viable option, although it is a linear model with a linear trend and/or seasonality. Traffic accidents and casualties by time-related qualities are crucial ideas to consider while analyzing the data based on these factors. Before making a choice, it is always a good idea to evaluate the performance of various models.

Target variable:
a. "Number_of_Casualties" - To predict the number of casualties in traffic accidents.
b. "Number_of_Vehicles" - To predict the number of vehicles involved in traffic accidents.

```
Call:
arima(x = train_data$Total_Casualties, order = c(best_params$p, best_params$d,
    best_params$q))

Coefficients:
         ar1
      -0.6364
s.e.   0.1845

sigma^2 estimated as 202299:  log likelihood = -113.17,  aic = 230.35

Training set error measures:
                   ME     RMSE      MAE      MPE     MAPE     MASE       ACF1
Training set -73.82354 435.4947 353.2006 175.2602 177.4992 0.7982536 -0.06732959
```

**Fig. 19** ARIMA Model

The output provides the best ARIMA model fitted with training data. Let's go through each part of the output:

**Call:** This section shows the function call used to create the ARIMA model. It indicates that the model was fitted using the best parameters found in the grid search.
**Coefficients:** This section shows the estimated coefficients for the AR (auto-regressive) part of the model. In your case, there is only one AR coefficient (ar1) with a value of -0.6364 and a standard error of 0.1845. The standard error provides an estimate of the uncertainty around the coefficient value. sigma^2 estimated as 202299: This line shows the estimated variance (sigma^2) of the residuals in the model. The log-likelihood and AIC (Akaike Information Criterion) values are also provided. The log-likelihood measure measures how well the model fits the data, while the AIC is a measure of model quality that balances the goodness of fit and model complexity. Lower AIC values indicate better models.
**The training set error measures:** This section provides various error metrics calculated on the training set. These metrics can be used to assess the model's performance:
ME (Mean Error): The average difference between the actual values and the predicted values. In your case, it is -73.82354.
**RMSE (Root Mean Squared Error):** The square root of the average squared difference between the actual values and the predicted values. In your case, it is 435.4947.
**MAE (Mean Absolute Error):** The average of the absolute differences between the actual values and the predicted values. In your case, it is 353.2006.
**MPE (Mean Percentage Error):** The average percentage difference between the actual values and the predicted values. In your case, it is 175.2602.
**MAPE (Mean Absolute Percentage Error):** The average of the absolute percentage differences between the actual values and the predicted values. In your case, it is 177.4992.
**MASE (Mean Absolute Scaled Error):** This is the mean absolute error of the model divided by the mean absolute error of a naïve forecast. A MASE value less than 1 indicates that the model is better than a naïve forecast. In your case, it is 0.7982536, which suggests that the model is better than a naïve forecast.
**ACF1 (Auto-correlation Function at Lag 1):** This measures the correlation between the model residuals at different lags. In your case, it is -0.06732959, which is close to 0, suggesting that there is little autocorrelation in the residuals.
Based on these results, the ARIMA model for forecasting and further analysis. The ARIMA model and have calculated a range of error metrics.

## 4.11 Justification For the Selection of Methods and Assumptions Made

Regression analysis was used to investigate the significant impact of various explanatory variables on the number of accidents on the roads. The choice of using the regression model is pegged on the fact that it is easier to get insights on how various variables impact the variable of interest, which is the response variable. Therefore, with regression analysis one can study the impact of variables on the dependent variables. The model takes the assumption that there is a linear relationship between the dependent variables and the explanatory variables. Secondly, they should be no relationship between the independent variables towards each and the error terms should be normally distributed

## 4.12 A Detailed Description of The Statistical Models and Tests Used

Basically, statistical models were used in the analysis and in the specific regression model was the best suited to derive insights from the Accident dataset. Our model is based on the explanatory variables and the response variables. Relationship between the variables was specifically what we needed to assess their impact. In addition, a correlation test was conducted between the variables of interest to investigate if there exists a relationship or correlation with each other.

### 4.12.1 Model Result Visualization by Performance Metrics:

Please see **Fig. 20** for more details.

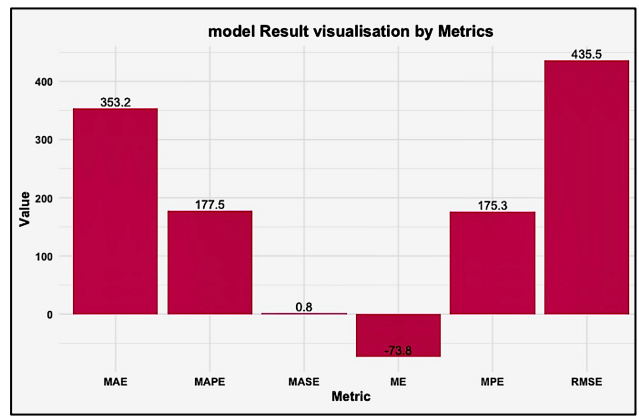

**Fig. 20** Model Result Visualisation by Performance Metrics

The model with the lowest error values (for metrics like RMSE, MAE, MPE, MAPE, and MASE) and the highest values for performance measures like the Log-likelihood would be considered the best model. Also, a lower AIC value is preferred as it indicates a model with a better fit. In this model, MASE of .800 shows the best result among them and ME of -73.80 shows a systematic bias in the model predictions which tends to underestimate the actual values

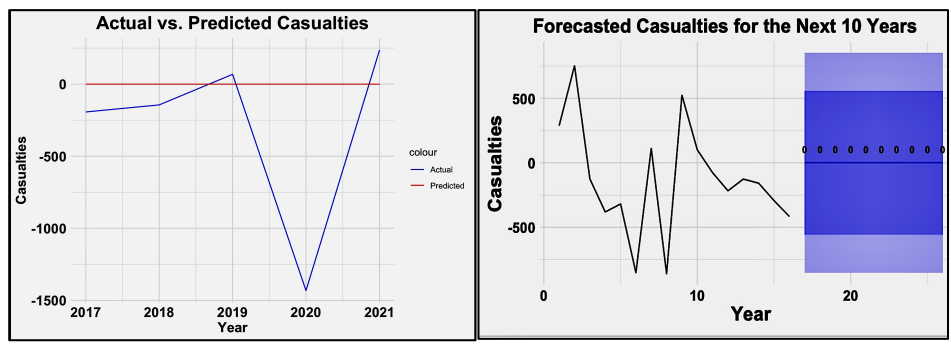

**Fig. 21** Actual Vs Predicted Casualties and Casualties Forecasting by The Year

**Actual vs. predicted values**: This step refers to comparing the actual and predicted values **Fig. 21** for the test se t. It helps assess the model's performance based on the data that it has not seen during the training process. This s tep is essential to ensure the model generalizes well to unseen data and is not overfitting the training data. **Forec** ast future values: This step refers to using the trained model to forecast values **Fig. 21** for future time periods that are not present in the dataset. This is the primary purpose of the time series analytical model to predict the behaviour of the series in the future based on historical data. The plot shows the next 10-year outcome as a prediction which is 0 level. predicted well and the number of casualties next 10 years would be lesser than the present number. Actual vs. predicted values are for evaluating the model's performance, while forecast future values are for using the model to make predictions for future time periods.

## 4.12.2 Econometric Analysis for Time Series:

Incorporating advanced statistical and econometric methods **Fig. 22** into time-series analysis is critical to comprehensively address the intricacies of real-world data. The ARIMA model is a univariate model, whereas the VAR model is a multivariate model. It means it is used to forecast a single time series. ARIMA excels in handling trends and seasonality in a single series, but it does not take into account the influence of other variables. Among these advanced techniques, the Vector Autoregressive (VAR) model stands out. It's a multivariate forecasting method, able to capture and predict the linear interdependencies among multiple time series. Moreover, VAR models can be further enhanced to become Structural VAR (SVAR) models, enabling causal interpretation of the relationships within the system. These methods provide superior sequence data prediction. Thus, the use of these methods can yield richer and more accurate insights from time-series data. By using the var model we got prediction casualties number and collisions.

```
$collisions_ts
           fcst       lower       upper         CI
  [1,]   152.6031   -12.95964   318.1659   165.5628
  [2,]   435.5057   240.41555   630.5958   195.0901
  [3,]   443.1474   247.17484   639.1200   195.9726
  [4,]  1063.1156   865.41938  1260.8118   197.6962
  [5,]  1445.4499  1247.18747  1643.7124   198.2625
  [6,]   142.5428   -96.31195   381.3975   238.8547
  [7,]   425.6735   172.94105   678.4060   252.7325
  [8,]   441.3046   188.34666   694.2625   252.9579
  [9,]  1007.6809   753.26038  1262.1014   254.4205
 [10,]  1388.5842  1133.55222  1643.6163   255.0320

$casualties_ts
           fcst       lower       upper         CI
  [1,]   132.6595  -109.89060   375.2097   242.5501
  [2,]   466.6136   177.46823   755.7590   289.1454
  [3,]   482.5977   192.50126   772.6942   290.0965
  [4,]  1450.2003  1159.09480  1741.3059   291.1055
  [5,]  1899.9077  1608.46039  2191.3550   291.4473
  [6,]   127.3235  -233.35642   488.0034   360.6799
  [7,]   460.8179    74.36842   847.2674   386.4495
  [8,]   480.7739    94.22943   867.3183   386.5544
  [9,]  1359.8011   971.45290  1748.1494   388.3482
 [10,]  1807.6734  1419.09200  2196.2549   388.5814
```

**Fig. 22** Econometric Analysis for Time Series

fcst: This is the forecasted value of the variable for that period. This is the best guess for what the value of the variable will be.
lower: This is the lower bound of the 95% confidence interval for the forecast. There is a 95% chance that the actual value will be higher than this number.
upper: This is the upper bound of the 95% confidence interval for the forecast. There is a 95% chance that the actual value will be lower than this number.
CI: This is the width of the 95% confidence interval. A wider interval suggests more uncertainty about the forecast.
The predictions for casualties_ts work in the same way as those for collisions_ts. In the first period, the model predicts collisions_ts to be approximately 152.60. However, there is uncertainty about this prediction. The model is 95% sure that the actual value will be between approximately -12.96 and 318.17.

## 5 Results And Interpretations

An important portion of a research study is where the results of the investigation are given and in-depth are explored (Chandler et al., 2019). The results of the investigation of the data gathered are presented in this chapter. Tables, graphs, and charts are frequently used to effectively communicate the findings and show the results in a clear and orderly way (Chandler et al., 2019). This information is presented in a targeted and organized manner since it is directly related to the study's goals or questions.

### 5.1 Key Findings

The mean age for drivers in accidents across all accident severities is between 37-50 and most accidents involve drivers aged between 25 and 65. The highest average age of a driver recorded as casualties were between the ages of 25-35. Lastly, 26-35 is the age band with the highest number of casualties which were fatal and 20-55 was the age band with more casualties.

The findings suggest that certain age groups are more prone to accidents and casualties, with the most vulnerable groups being drivers aged 25-65. More specifically, drivers between 26-35 years old are associated with the highest number of fatalities, while the 20-55 age band reports the most casualties. This information is invaluable in guiding road safety policy and practice.

Targeted Education and Training Programs: One possible implication is that road safety initiatives should be tailored to these at-risk age groups. Drivers in the age group of 26-35, for instance, might benefit from education and training programs that focus on safe driving practices, the consequences of risky behaviour, and the importance of factors like seatbelt use, sober driving, and following speed limits.

Media Campaigns: Media campaigns designed to target these specific age groups could also be beneficial. These could include advertisements on social media platforms, television, radio, and other popular media channels for these age groups.

Vehicle Safety Technology: The findings might also inform initiatives aimed at implementing and promoting safety technology in vehicles. Younger drivers might benefit from technology such as collision detection systems, automatic braking, and other advanced driver-assistance systems (ADAS).

Insurance Incentives: Insurance companies could be encouraged to offer incentives for drivers in these age groups who take part in advanced driving courses or use safety-enhancing technologies in their vehicles.

Infrastructure Improvements: The findings could also guide infrastructure improvements. For example, the high casualty rate among the 20-55 age group might suggest a need for safer road designs in areas where these age groups are most likely to drive.

Workplace Policies: Given the high mean age of drivers in accidents, companies may need to consider safe-driving policies and regular training for employees who drive as part of their work, especially those in the 37-50 age bracket. Findings offer useful insights that could help inform a multi-faceted approach to road safety. The study is dependent on readily available data sources, and this is one of the limitations in terms of assessing the severity of traffic incidents in the UK. The quality, completeness, and correctness of the submitted data are also a constraint on the study's conclusions and analysis. Inaccuracies or biases are introduced into the results when the methods used for data collecting or reporting were incorrect. The study also only includes reported events, leaving out unreported or minor incidents that could nevertheless offer important information about the severity of an accident.

## 5.2 Number of Casualties by ( Age Band of Driver and Accident Severity)

## 5.3 Count vs Accident Occurrence

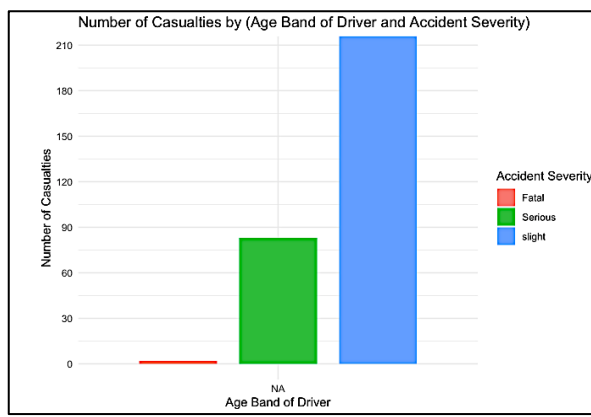
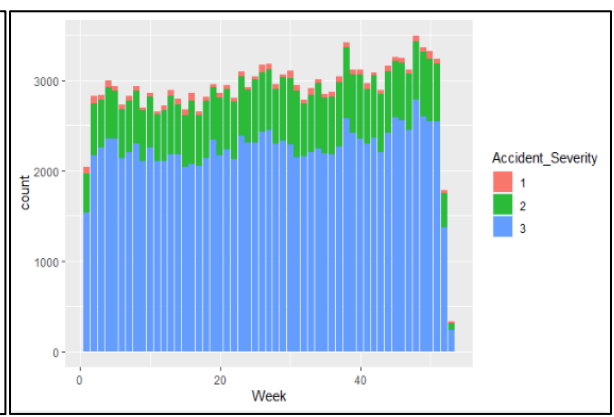

Fig. 23 Number of Casualties by Age Band     Fig. 24 Count vs Accident Occurrence

According to **Fig. 23** age bad slight age band is showing a higher number of casualties than other categories which are near 215. According to **Fig. 24** , the initial count for week 1 and the last few weeks of the year have the lowest accident counts- perhaps linked to holiday seasons and most people spending time with loved ones, so not being out in traffic much. There isn't a specific trend we can pinpoint, other than week 47 having the highest count.

## 5.4 Distributions Of Age and Accident Severity

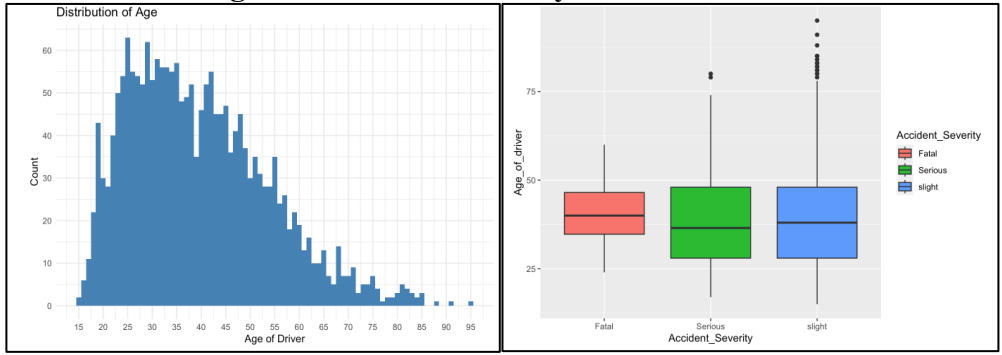

      **Fig. 25** Distribution of Ages       **Fig. 26** Boxplot showing the relationship between age and accident severity

From the above **Fig. 25**, it is clear the highest average age of a driver recorded as casualties were between the ages of 25-35. This meant that most of the casualties were youths and adults. Referring to **Fig. 26**, it's clear that the severity of accidents is highest for the 'serious' category compared to 'fatal' and 'slight'. When considering age, the average age for fatal accidents is 46, while for slight accidents it is 48. Interestingly, the age group most associated with serious accidents is 49, which is higher than both the other categories.

## 5.5 Distributions Of Week and Number of Casualties By Weather Conditions

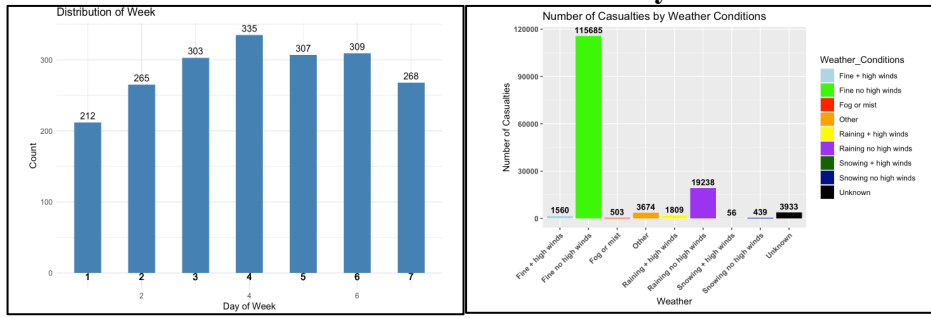

      **Fig. 27** Distribution of weeks.       **Fig. 28** Weather Conditions

According to the chart **Fig. 27**, week 4 shows the highest number distribution of the week whereas week 1 lowest According to the chart **Fig. 28**, the number of casualties showed in fine no high winds. And snowing high winds very low number of casualties. 115685 number of casualties fine no high winds and snowing and high winds for 56. Rest of weather conditions for the number of casualties respectively.

## 5.6 Distribution Of Number Of Vehicles And Speed Limits In Mph

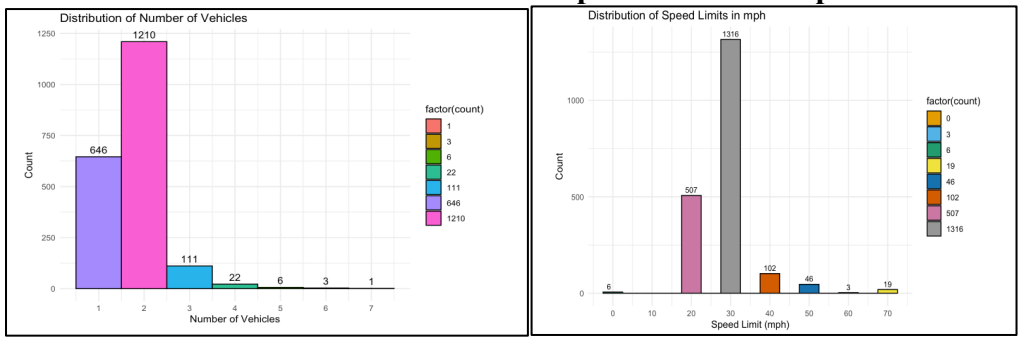

      **Fig. 29** Number of Vehicles       **Fig. 30** Distribution of Speed Limits in Mph

The bar graph **Fig. 29** shows the distribution of the number of vehicles in a sample dataset. The first column represents the count of vehicles, and the second column represents the frequency of each count. For instance, there was 1 vehicle with a count of 1210, 3 vehicles with a count of 646, 111 and 22 respectively and so on. **Fig. 30** shows the distribution of speed limits in mph. We can see that the possible speed limit values are 0, 10, 20, 30, 40, 50, 60, and 70 mph. The count column represents the number of times each speed limit value appears in the data. For example, the speed limit of 40 mph appears 102 times in the data, while the speed limit of 70 mph appears 19 times. The highest count is 1316, which corresponds to the speed limit of 30 mph.

## 5.7 Number of Casualties by Light Conditions and Road Type

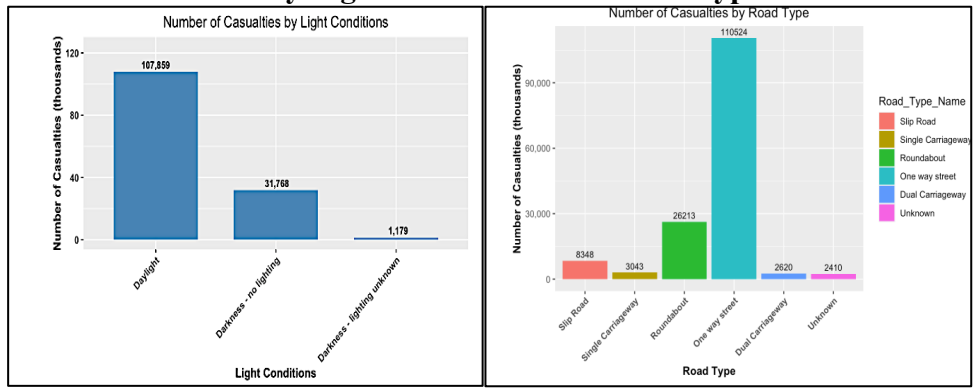

**Fig. 31** Number of Casualties by Light Conditions  **Fig. 32** Number of Casualties by Road Type

**Fig. 31** shows the number of casualties by light conditions, with the number of casualties in thousands on the vertical axis and the different light conditions on the horizontal axis. It seems to include four categories: daylight, darkness with no lighting, darkness with unknown lighting, and other light conditions. There were 107,859 casualties in daylight conditions and 31,768 casualties in darkness with no lighting. The number of casualties in darkness with unknown lighting is noted as 1179, and the number of casualties in other light conditions is noted as about 118. More dangerous conditions have fewer casualties, which is different to initial assumptions. **Fig. 32 above** shows the number of casualties by road type. The table shows the number of casualties in thousands for different road types including Slip Road, Single Carriageway, Roundabout, One-way Street, Dual Carriageway, and Unknown. The highest number of casualties seems to have occurred in a one-way street with 110524 casualties, followed by a roundabout with 26213 casualties. Slip Road and single carriageway seem to have the lowest number of casualties with 8348 and 3043 respectively

## 5.8 Number of Casualties by Special Condition and Road Surface Condition

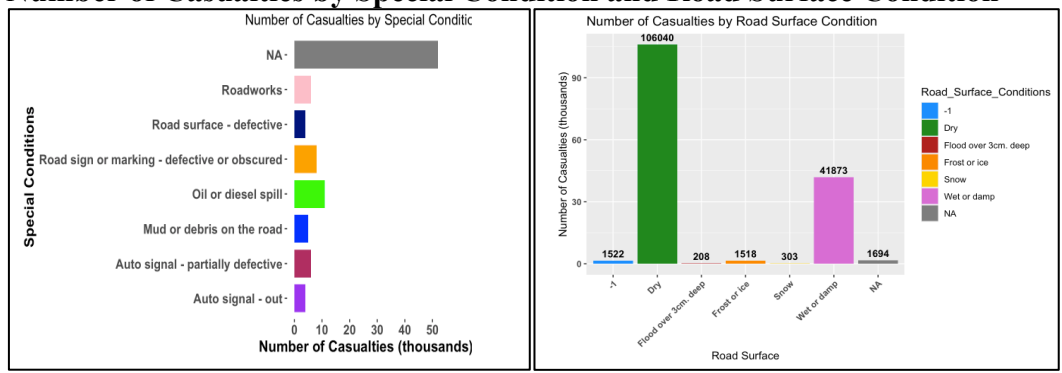

**Fig. 33** Number of Casualties by Special Condition. **Fig. 34** Number of Casualties by Road Surface Condition

**Fig. 33** provides information on the number of casualties (in thousands) caused by road accidents under specific special conditions that may contribute to the occurrence of the accident. The special conditions include "NA," "Roadworks," "Road surface - defective," "Road sign or marking - defective or obscured," "Oil or diesel spill," "Mud or debris on the road," "Auto signal - partially defective" and "Auto signal - out." Each of these conditions has a corresponding bar chart representing the number of casualties. The height of each bar represents the number of casualties (in thousands) for that specific condition. On the other hand, the lowest number of casualties occurred under the special condition of "oil and diesel spill with a value of approximately 10,000. Therefore, the chart suggests that special conditions on the road pose a significant risk to drivers' and passengers' safety, and measures should be put in place to reduce the risk of accidents occurring under these conditions. **Fig. 34** provides information on the number of casualties (in thousands) by Road Surface Condition. It displays a total of six types of Road Surface Conditions: Dry, Flood over 3cm. deep, Frost or ice, Snow, Wet or damp, and NA, which denotes missing values. The first highest number of casuals indicates that there were 106,040 casualties in total. Dry road surface conditions had the highest number of casualties, while Snow had the lowest number of casualties. For example, 1522 casualties were recorded for unknown surface conditions, while 208 casualties were recorded for flood over 3cm deep road surface conditions. Taking data from the government website about regions and their population, we performed analysis trying to see if we can spot trends in relation to accidents. In the chart, in wet or damp conditions 41873 casualties show.

## 5.9 Number of Accidents by Percentage in the UK top-10 place:

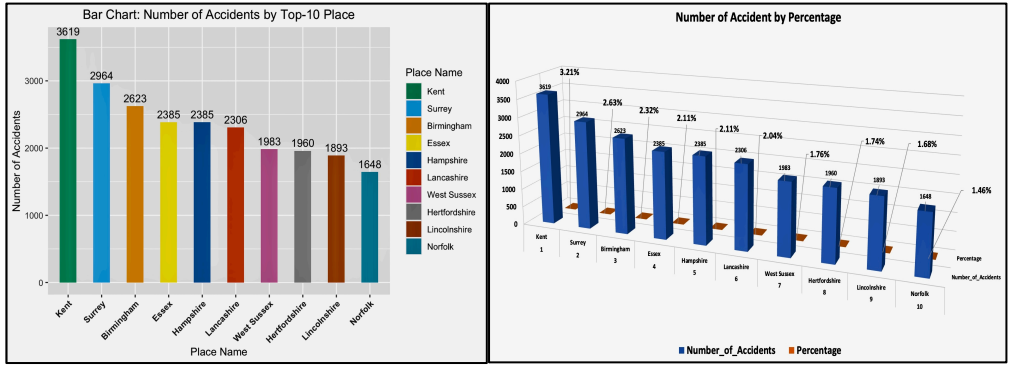

Fig. 35 Number of Accidents in the UK top-10 place    Fig. 36 Number of Accidents by Percentage

According to the bar chart **Fig.35** there are showing Kent area's number of accidents is higher whereas Norfolk is lesser compared to other places. So, the number of accidents is 3619 in Kent whereas 1648 is in the Norfolk area. 2964 in Surrey whereas 1893 in Lincolnshire in the United Kingdom. According to the percentage **Fig.36 above** Kent area is 3.21 per cent whereas the loosest percentage is 1.46in Norfolk in the United Kingdom.

## 5.10 Number of Collisions & Casualties per Year

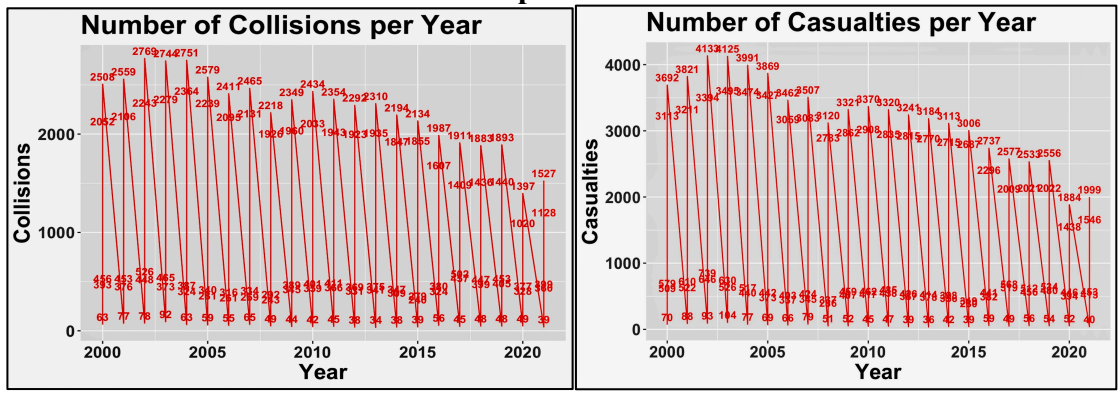

Fig. 37 Number of Collisions per Year    Fig. 38 Number of Casualties Per Year

The provided figure illustrates a twenty-year trend in the number of collisions per year **Fig.37**, spanning from 2000 to 2020. In 2000, collisions were notably high, with a total of 2,508 reported cases. This figure rose in the subsequent two years, peaking at 2,769 collisions in 2002. Interestingly, an increase in collisions typically corresponds with a rise in accident-related casualties. By 2005, the number of collisions had slightly decreased, approximating the figure from the year 2000. The decade starting from 2010, however, marked a significant reduction in the annual number of collisions. Five years later, in 2015, the reported collisions dropped further to 2,134. This figure is substantially lower than the numbers recorded in the preceding decade. By 2020, the total number of collisions had decreased drastically to 1,128. This represents a significant shift in the trend over the twenty-year period under review. The given figure **Fig. 38** presents a twenty-year trend in the number of casualties per accident from 2000 to 2020. In the year 2000, the casualties were notably high, standing at 3,692. This number escalated over the next two years, reaching a peak of 4,134 casualties in 2002. The subsequent five years saw significant changes in the casualty rate, but the trend generally remained high. However, a noticeable decrease in the number of casualties began to emerge around 2010, initiating a downward trend that continued for the next decade. By 2015, the casualty rate had dropped considerably compared to the turn of the century. This downward trend continued, and by 2020, the total number of casualties was recorded as 1,546. This figure represents a significant reduction in casualties compared to the statistics from fifteen years prior.

## 5.11 Explainable AI (XAI)
### 5.11.1 SHAP Model of Feature Importance: Fig. 39

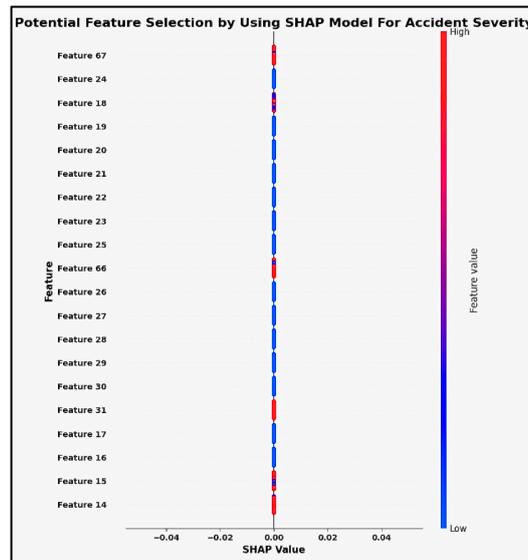

**Fig. 39** SHAP Model of Feature Importance by using SHAP Models for Accident Severity

Explainable AI (XAI) is important because it aims to create models that humans can easily understand. This transparency allows individuals to comprehend the reasons behind the decisions or predictions made by AI models. The SHAP (Shapley Additive exPlanations) model is an XAI technique that provides explanations for the output of any machine learning model. It uses cooperative game theory to determine the contribution of each feature in predicting the outcome for individual instances. This comprehensive approach captures feature interactions and helps identify the importance of each feature. According to the SHAP analysis conducted in our research, the following features were found to have significant impacts on accident severity:
index in [67, 18, 66, 31, 14, 15]] of Driver_Home_Area_Type, Longitude, Driver_IMD_Decile, Road_Type, Casualty_Home_Area_Type, and Casualty_IMD_Decile respectively. The insights gained from the SHAP analysis shed light on the specific features that play a crucial role in predicting accident severity. For that observation, the hue indicates whether that variable is high (in red) or low (in blue). By understanding the importance of these features, we can develop targeted interventions and policies to improve road safety. For example, addressing factors related to driver home area type, longitude, driver IMD decile, road type, casualty home area type, and casualty IMD decile can contribute to more effective strategies in reducing accident severity. Explainable AI allows for a better understanding of AI model decisions, and the SHAP model specifically helps analyse feature importance in predicting accident severity. By identifying influential features, we gain valuable insights that can inform interventions and policies for enhancing road safety and lead to effective strategies for improving road safety.

## 5.12 Random Forest Classifier Model [35][36]
We had to define the potential feature among those variables and the target variable mean predicted dependent was accident severity.

```
features = merged_df[['Speed_limit', 'Weather_Conditions',
'Light_Conditions', 'Road_Surface_Conditions', 'Vehicle_Type',
'Age_of_Driver']]
target = merged_df['Accident_Severity']
```

The mathematical model of the random forest classifier **Fig. 40** can be represented as follows:

$$P(y = c \mid x) = 1/n \sum_{i=1}^{n} I(y_i = c) \tag{1}$$

$P(y=c|x)$ is the probability that a new data point x belongs to class c.
n is the number of trees in the forest.

$I(y_i=c)$ is an indicator function that is equal to 1 if $y_i=c$ and 0 otherwise.
The sum $\Sigma$ goes from $i = 1$ to n.
The significance of Random Forests includes:
1. Accuracy: Random Forests are often very accurate, performing well in many scenarios compared to other algorithms.
2. Avoids overfitting: By averaging out the results of many trees, Random Forests help to avoid overfitting, a common problem with decision trees.
3. Feature importance: Random Forests offer a great way to estimate feature importance.
4. Robustness: They are robust to outliers and the distribution of the data, making them very versatile.
5. Ease of use: They generally work well out of the box, and don't require a lot of tuning.

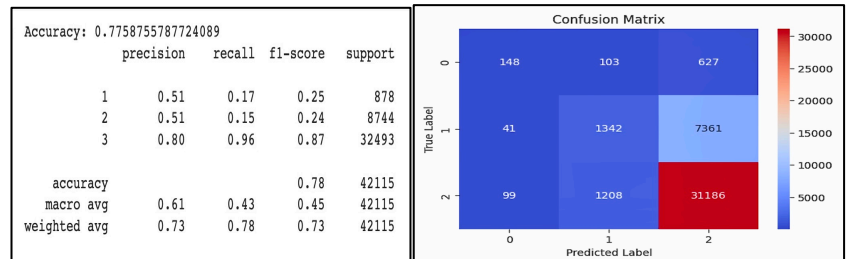

**Fig. 40** Random Forest Classifier Model Accuracy Visualisation by Confusion Metrics

The random forest classifier showed 77 % overall accuracy and a precision score of 73 percent recall 78 % F1-Score 73 %

## 5.13 H20 autoML processing to make the best machine learning model:[34]

H2O's AutoML is an automated machine-learning process. It automates many of the repetitive tasks involved in machine learning, such as data pre-processing, feature selection, model selection, and hyperparameter tuning, to make machine learning more accessible and efficient. The goal of the H2O AutoML process is to produce a "leaderboard" of the best models it found, ranked by their performance on the validation data. Rather, it's a tool or process that helps in developing machine learning models. From the H2O AutoML execution, the best model chosen is the XGBoost model with a model key `XGBoost_1_AutoML_6_20230718_230700`. This model trained 345 trees for prediction.

The final result is a table (referred to as a leaderboard) that provides the metrics of the best-performing model. From the results, we can see that the best model was an XGBoost model. XGBoost stands for eXtreme Gradient Boosting, which is a highly efficient and powerful tree-based machine learning algorithm. The AutoML process has helped us to identify a powerful XGBoost model that appears to predict your target variable (Accident_Severity) quite well, judging by the metrics. It is a good idea what models perform well on the dataset without having to individually train and optimize each model.

The target variable, Accident Severity, directly addresses the objectives. It provides insights into how driver age, environmental and road conditions, and driver distractions might influence the severity of accidents. Identifying High-Risk Groups: Based on the predictive analysis, it can identify the groups (e.g., age groups, seasons, road conditions etc.) that are more likely to be involved in severe accidents. This directly relates to objectives 1, 2, and 3.

Model Performance:

RMSE (Root Mean Squared Error): This measures the average squared difference between the predicted and actual values. The square root of these means is then taken to bring the errors back to the same unit as the original target. In this case, the RMSE is 0.172834. Lower RMSE values indicate a better fit to the data.

MSE (Mean Squared Error): This is the average squared difference between the predicted and actual values. It's basically the square of RMSE and in your case, it's 0.0298717.

MAE (Mean Absolute Error): This measures the average absolute difference between predicted and actual values. It's less sensitive to outliers than the MSE. In this case, the MAE is 0.0871839.

RMSLE (Root Mean Squared Logarithmic Error): This is a variation of RMSE that is calculated on the logarithm of the predicted and the actual values. It's mainly used when the target variable has a wide range of values to prevent the model from being overly influenced by large values. In this case, the RMSLE is 0.0540801.

Mean Residual Deviance: In general, the lower this value, the better the model's predictions match the actual values. In your case, it's 0.0298717.

The model informs policies and interventions to reduce accident severity, which aligns with the overall aim. For example, a strong relationship between accident severity and driver age or specific distractions could be the basis for targeted interventions or public awareness campaigns. recommendations for reducing the frequency and severity of road traffic accidents, as specified in objective 6.

ModelMetricsRegression: XGboost **Fig. 41**

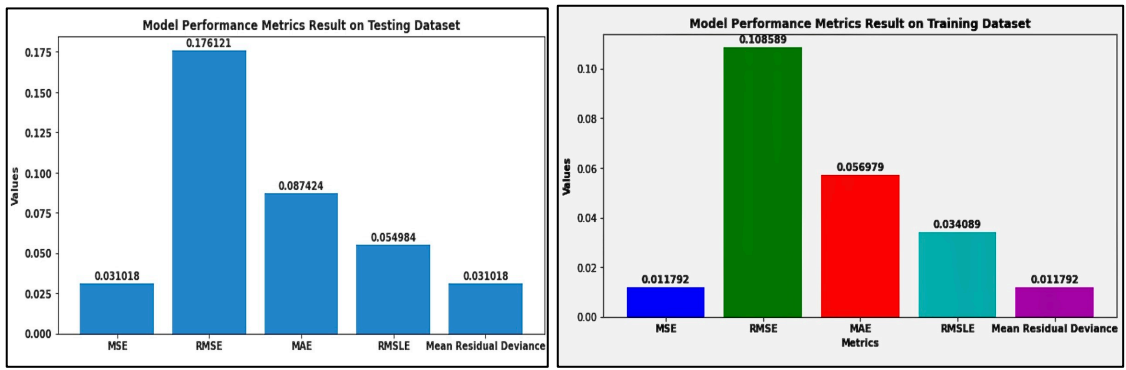

**Fig. 41** Best Model Performance Metrics of XGBoost on Training and Testing Dataset

Based on these results, the XGBoost model appears to have reasonable predictive performance for accident severity.

## 5.13.1 Scoring History:[37][38]

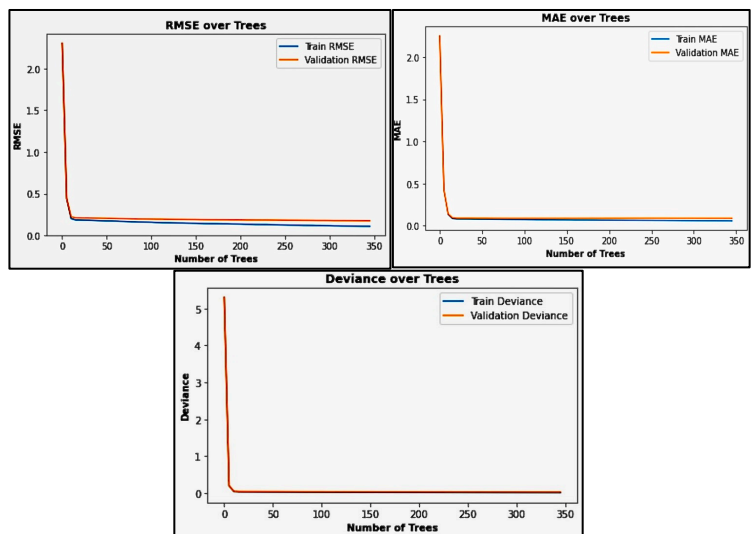

**Fig. 42** Scoring History of Best Model

The scoring history **Fig. 42** helps us understand how our model is learning. For instance, if we see that the validation error starts increasing while the training error is still decreasing, this might be a sign of overfitting. It records the model's performance on certain metrics at each iteration or epoch of the training process. The model training began at "2023-07-18 23:07:02" and ended at "2023-07-18 23:16:58", which means the total training time for this model was approximately 10 minutes. The initial model (with 0 trees) was created almost instantaneously as the duration was logged as "0.013 sec". As the number of trees increased, so did the duration, indicating that more complex models took longer to train. By the time 5 trees were trained, 21.485 seconds had passed. It took just over a minute (1 min 1.814 sec) to train 40 trees. The model continued to train more trees, reaching 300 trees in approximately 8 minutes (8 min 2.681 sec). The final model, with 345 trees, finished training in approximately 10 minutes (9 min 56.783 sec).

# 6 Discussion

## 6.1 Comparison Of the Findings with Previous Research and Theoretical Expectations

From the findings, it is assessed that most youth or persons between the age of 25-35 are the most individuals at high risk of being among the road accident casualties. The cause of these accidents is registered to be associated with less weather and lighting conditions as could be anticipated. From descriptive statistics, the average number of causalities in road accidents in the United Kingdom is (1.196). The average number of vehicles involved in road accidents is (2). The average number of accidents recorded within a week is (4) while the average age of a person involved in a road accident is approximately (40) years old. This explains that the major people causing an increase in the number of accidents in the United Kingdom are the youths. The R squared of the multiple regression model containing our variables of interest was at 2% which implies that most of the factors influencing the number of casualties recorded in the road accidents were outside the model, they were not captured within the data model. The coefficients of the speed limit and the number of casualties were statistically significant since their p-values were less than 0.05, indicating they were significant predictors of the number of casualties recorded in the road accident in the United Kingdom. The coefficients had positive signs showing there was a positive relationship between the number of casualties level and the speed limit. Therefore, an increase in speed limit level increases the number of accident pressure. To encourage economic progress and the eradication of poverty in underdeveloped but democratic countries, the United States government supports the Millennium Challenge Corporation (MCC). According to the researcher **[8]**, the legislation requires MCC to commit the full grant money at a project's commencement and complete it within five years. Hence, cost variability has an impact on MCC transportation infrastructure projects. There is a shortage of quantitative literature that focuses on the construction stage of transport infrastructure costs in developing nations. In this study, mitigation methods were found and the cost variability for MCC road construction projects was assessed. The average increase from the financial authority to the actual expenses was 135%. The design phase of the project had the highest degree of uncertainty about cost estimates; there was a mean 100% rise between financial authorization and the engineer's estimations. o curbs the high incidence of road accidents among the youth, more comprehensive driver education programs and stricter regulations for younger drivers may be necessary. As for the significant variability in costs for MCC infrastructure projects, a more rigorous and accurate cost estimation method during the design phase could help reduce this uncertainty. However, it is important to note that this study had certain limitations. The factors contributing to road accidents are multifarious, and our model could capture only a fraction of these. Additionally, the complex nature of infrastructure projects means that cost overruns are often inevitable. It is also noteworthy to mention the seemingly disjointed nature of the two sections presented here. While road safety and infrastructure development may seem distinct, they are in fact interconnected in the larger scope of public safety and economic development. Therefore, improvements in one area can have positive repercussions on the other."

## 6.2 Implications of the Findings for Road Safety Policies and Practices

There are several findings with several consequences for practices and policies related to traffic safety. These implications also contain recommendations for further study and solutions.
**Good Road Design:** To lower the risk of accidents, road safety measures should concentrate on constructing safer roadways. This includes enhancing the road network by widening roadways and installing security barriers.
**Speed Reduction Measures:** To lower the risk of accidents, policies should concentrate on lowering speed limits and putting in place speed reduction measures, like speed cameras and traffic measures.
Road safety can be promoted by education and awareness programs, which are important. The policy should emphasize promoting safe driving habits and educating motorists, passengers, and pedestrians on the dangers of imprudent behaviour on the road.
**Enforcing Current Laws:** The policy should concentrate heavily on upholding current road safety regulations, such as seatbelt requirements, and cracking down on unsafe driving practices, like operating a vehicle while under the influence of drugs or alcohol.
**Invest in Research:** Future research should focus on discovering innovative approaches and tools to enhance traffic safety, such as bettering driver assistance technology and enhancing car safety features.

## 6.3 Recommendations For Interventions and Future Research

Road safety policies and procedures should focus on creating safer roads, lowering speed limits, educating the public about the dangers of risky behaviour, upholding the law, and funding research to find innovative methods and technology to enhance road safety.

## 6.4 Examination of the Limitations and Strengths of the Study

**Strengths:**

*Comprehensive Data Analysis:* The study utilizes extensive data from the UK government, which provides a rich basis for understanding the factors influencing the severity of road traffic accidents.
*Use of Multivariate Analysis:* The application of multivariate statistical analysis enables the examination of complex interactions among variables, enhancing the comprehensiveness and robustness of the findings.
*Focus on Casualty Status 19:* The specific focus on Casualty Status 19 allows for a nuanced understanding of accident severity, contributing unique insights to the literature on road safety.

**Limitations:**

*Limited Scope:* The study exclusively concentrates on UK road traffic accidents in 2019, potentially limiting the generalizability of its results to different locations or time periods.

*Data Quality and Validity:* The study's findings are heavily reliant on the accuracy and completeness of the UK government-collected data. Any errors or omissions in this dataset may affect the validity of the conclusions drawn.

*Variable Selection:* The study utilizes a select number of variables to evaluate the determinants of accident severity. Other potentially impactful factors, such as weather conditions or road status, are not considered.

*Statistical Techniques:* While the study employs commonly used statistical techniques, like regression analysis, it is unclear if all the necessary assumptions for these methods were verified for ensuring reliable results.
*Machine learning model*: building time H2O autoML took a long time to execute which was a limitation for machines such as model building.

## 6.5 Discussion of the Broader Implications of The Research for Public Health And Society.

One important finding from the research is that it outlines the most important causes of RTA severity in the UK. It was discovered that discovered factors such as age, gender, vehicle type, speed limit, and meteorological conditions were important predictors of RTA severity. These results can help guide the policies promoting road safety, such as establishing suitable speed limits, raising vehicle safety requirements, and improving weather predicting abilities. The study's machine learning and advanced statistical data analysis approach is also crucial for society and public health. The dynamic interactions between different factors and how they affect the severity of RTA are better understood through multivariate analysis. This method can assist decision-makers in creating focused interventions that address the most important causes of the severity of RTA, resulting in better public health outcomes. The study also highlights the importance of data gathering and analysis in raising traffic safety, which is another implication. Any inaccuracies or missing data could compromise the validity of the findings because the study depended on the accuracy and comprehensiveness of the data gathered by the UK government. This highlights the need for effective data collection and analysis mechanisms to support evidence-based policies and initiatives to enhance road safety. Another angle to consider is the impact of these findings on specific populations in society as well as public health. Certain groups might be more vulnerable to RTAs, such as the elderly or young drivers, warranting special attention in road safety initiatives. Furthermore, our research highlights the importance of addressing behavioural elements in road safety. Factors such as speed limit and vehicle type hint at the role of personal choices and behaviours in the severity of RTAs, suggesting that strategies for promoting safer driving habits could be beneficial. Lastly, while this study has shed light on several key predictors of RTA severity, it also underscores the need for further research. Some variables were outside our model, highlighting the necessity for continued exploration of other potential factors influencing RTA outcomes.

## 7 Conclusion

This research project explored UK road traffic accident severity employing an enhanced array of techniques including machine learning, econometric techniques, and time series forecasting. Leveraging 2019 Road Safety Data and longitudinal data from 1998 to 2019, we applied comprehensive approaches such as descriptive, inferential, bivariate, and multivariate methods, correlation analysis, regression models, the Generalized Method of Moments (GMM), and ARIMA time series forecasting. We also utilized the Random Forest classifier machine learning model, resulting in significant predictive accuracy, and Explainable AI (XAI) techniques such as the

SHAP model to gain better insights into the influential factors. Our findings demonstrated that driver age, seasonal trends, environmental and road conditions, number of vehicles involved, and driver distractions significantly affect accident severity. Drivers aged 25-65, especially those between 26-35, emerged as particularly vulnerable to severe accidents. These results inform targeted interventions, including education, media campaigns, vehicle safety technology, and insurance incentives, aimed at reducing accident frequency and severity. Infrastructure improvements in areas frequented by vulnerable age groups and workplace policies promoting safe driving can enhance road safety. It's important to acknowledge the limitations of this study, including reliance on readily available data sources, which might limit the quality and completeness of the analysis. Potential inaccuracies or biases in data collection or reporting methods could exist. Additionally, the study focuses on reported incidents, potentially excluding unreported or minor occurrences that could provide valuable insights. Future research should consider incorporating additional data sources, exploring non-linear relationships between variables, and investigating the impact of other factors. Utilizing advanced statistical techniques like time series analysis and AI-powered models can yield more accurate forecasts and facilitate evidence-based road safety policies and interventions. This study extends our understanding of the complex factors affecting UK traffic accident severity. The integrated use of AI technologies, machine learning, econometric techniques, and other statistical tools offers significant promise in informing evidence-based interventions and policies to improve road safety. Implementing the insights gained from this research could greatly enhance public health and safety, reducing the financial and human costs associated with traffic accidents.